%% file: main.tex
\documentclass[twoside]{article}

\usepackage[preprint]{aistats2026}

\usepackage{microtype}
\usepackage{graphicx}
\usepackage{booktabs} %
\usepackage{hyperref}
\usepackage{subcaption}
\usepackage[clockwise,figuresleft]{rotating}
\input{sections/0.topmatter}

\usepackage{amsmath}
\usepackage{amssymb}
\usepackage{mathtools}
\usepackage{amsthm}
\usepackage[capitalize,noabbrev]{cleveref}
\theoremstyle{plain}

\theoremstyle{definition}

\theoremstyle{remark}

\usepackage[textsize=tiny]{todonotes}

\begin{document}

\runningauthor{S. T. Truong, D. Q. Nguyen, W. Neiswanger, R-R Griffiths, S. Ermon, N. Haber, S. Koyejo}

\twocolumn[

\aistatstitle{Neural Nonmyopic Bayesian Optimization in Dynamic Cost Settings}

\aistatsauthor{Sang T. Truong$^1$, Duc Q. Nguyen$^{2}$, Willie Neiswanger$^3$, Ryan-Rhys Griffiths$^4$,\\ \textbf{Stefano Ermon$^1$, Nick Haber$^1$, Sanmi Koyejo$^1$}}

\aistatsaddress{ $^1$Stanford University, $^2$National University of Singapore,\\$^3$University of Southern California, $^4$FutureHouse, Inc. } ]

\input{sections/1.abstract}
\input{sections/2.introduction}
\input{sections/3.cost_taxonomy}
\input{sections/4.method}
\input{sections/5.experiment}
\input{sections/6.discussion}

\bibliography{sections/7.references}

\section*{Checklist}

\begin{enumerate}

  \item For all models and algorithms presented, check if you include:
  \begin{enumerate}
    \item A clear description of the mathematical setting, assumptions, algorithm, and/or model. [Yes. Section 3]
    \item An analysis of the properties and complexity (time, space, sample size) of any algorithm. [Yes. Appendix C]
    \item (Optional) Anonymized source code, with specification of all dependencies, including external libraries. [Yes. Abstract]
  \end{enumerate}

  \item For any theoretical claim, check if you include:
  \begin{enumerate}
    \item Statements of the full set of assumptions of all theoretical results. [Yes. Section 3]
    \item Complete proofs of all theoretical results. [Yes. Appendix C]
    \item Clear explanations of any assumptions. [Yes. Section 1, 3]     
  \end{enumerate}

  \item For all figures and tables that present empirical results, check if you include:
  \begin{enumerate}
    \item The code, data, and instructions needed to reproduce the main experimental results (either in the supplemental material or as a URL). [Yes. Abstract]
    \item All the training details (e.g., data splits, hyperparameters, how they were chosen). [Yes. Section 4, Appendix F, G]
    \item A clear definition of the specific measure or statistics and error bars (e.g., with respect to the random seed after running experiments multiple times). [Yes. Section 4, Appendix E, F]
    \item A description of the computing infrastructure used. (e.g., type of GPUs, internal cluster, or cloud provider). [Yes. Section 4]
  \end{enumerate}

  \item If you are using existing assets (e.g., code, data, models) or curating/releasing new assets, check if you include:
  \begin{enumerate}
    \item Citations of the creator If your work uses existing assets. [Yes. Section 4.2]
    \item The license information of the assets, if applicable. [Not Applicable]
    \item New assets either in the supplemental material or as a URL, if applicable. [Yes. Abstract]
    \item Information about consent from data providers/curators. [Not Applicable]
    \item Discussion of sensible content if applicable, e.g., personally identifiable information or offensive content. [Not Applicable]
  \end{enumerate}

  \item If you used crowdsourcing or conducted research with human subjects, check if you include:
  \begin{enumerate}
    \item The full text of instructions given to participants and screenshots. [Not Applicable]
    \item Descriptions of potential participant risks, with links to Institutional Review Board (IRB) approvals if applicable. [Not Applicable]
    \item The estimated hourly wage paid to participants and the total amount spent on participant compensation. [Not Applicable]
  \end{enumerate}

\end{enumerate}

\clearpage
\appendix
\thispagestyle{empty}

\onecolumn
\aistatstitle{Supplementary Materials}
\input{sections/8.appendices}
\input{sections/9.impact}

\end{document}

%% file: sections/0.topmatter.tex
\usepackage{amsmath,amsfonts,amssymb}
\usepackage{booktabs}
\usepackage{enumitem}
\usepackage{tikz}
\usetikzlibrary{positioning}
\usepackage{array}
\usepackage{url}
\usepackage{hyperref}
\usepackage{seqsplit}
\usepackage{listings}
\usepackage{graphicx}
\usepackage{multirow}
\usepackage{float}

\usepackage{tcolorbox}
\definecolor{lightblue}{HTML}{ff7f2a}
\definecolor{lighterblue}{HTML}{ffe6d5}%
\newtcolorbox{mybox}{colback=lighterblue,colframe=lightblue}

\newcommand{\Ac}{{\mathcal{A}}}

\newcommand{\Nc}{{\mathcal{N}}}
\newcommand{\Oc}{{\mathcal{O}}}

\newcommand{\Xc}{{\mathcal{X}}}
\newcommand{\Yc}{{\mathcal{Y}}}

\newcommand{\Cbb}{{\mathbb{C}}}

\newcommand{\Ebb}{{\mathbb{E}}}

\newcommand{\Hbb}{{\mathbb{H}}}
\newcommand{\Ibb}{{\mathbb{I}}}

\newcommand{\Rbb}{{\mathbb{R}}}

\DeclareMathOperator*{\argsup}{arg\,sup}
\DeclareMathOperator*{\arginf}{arg\,inf}
\DeclareMathOperator*{\argmax}{arg\,max}

\newcommand*\diff{\mathop{}\!\mathrm{d}}

%% file: sections/1.abstract.tex
\begin{abstract}

Bayesian optimization (BO) is a common framework for optimizing black-box functions, yet most existing methods assume static query costs and rely on myopic acquisition strategies. We introduce LookaHES, a nonmyopic BO framework designed for dynamic, history-dependent cost environments, where evaluation costs vary with prior actions, such as travel distance in spatial tasks or edit distance in sequence design. LookaHES combines a multi-step variant of $H$-Entropy Search with pathwise sampling and neural policy optimization, enabling long-horizon planning beyond twenty steps without the exponential complexity of existing nonmyopic methods. The key innovation is the integration of neural policies, including large language models, to effectively navigate structured, combinatorial action spaces such as protein sequences. These policies amortize lookahead planning and can be integrated with domain-specific constraints during rollout. Empirically, LookaHES outperforms strong myopic and nonmyopic baselines across nine synthetic benchmarks from two to eight dimensions and two real-world tasks: geospatial optimization using NASA night-light imagery and protein sequence design with constrained token-level edits. In short, LookaHES provides a general, scalable, and cost-aware solution for robust long-horizon optimization in complex decision spaces, which makes it a useful tool for researchers in machine learning, statistics, and applied domains. 
Our implementation is available at 
\href{https://github.com/sangttruong/nonmyopia}{https://github.com/sangttruong/nonmyopia}.
\end{abstract}

%% file: sections/2.introduction.tex
\begin{figure*}[hbt!]
    \centering
    \includegraphics[width=0.7\linewidth]{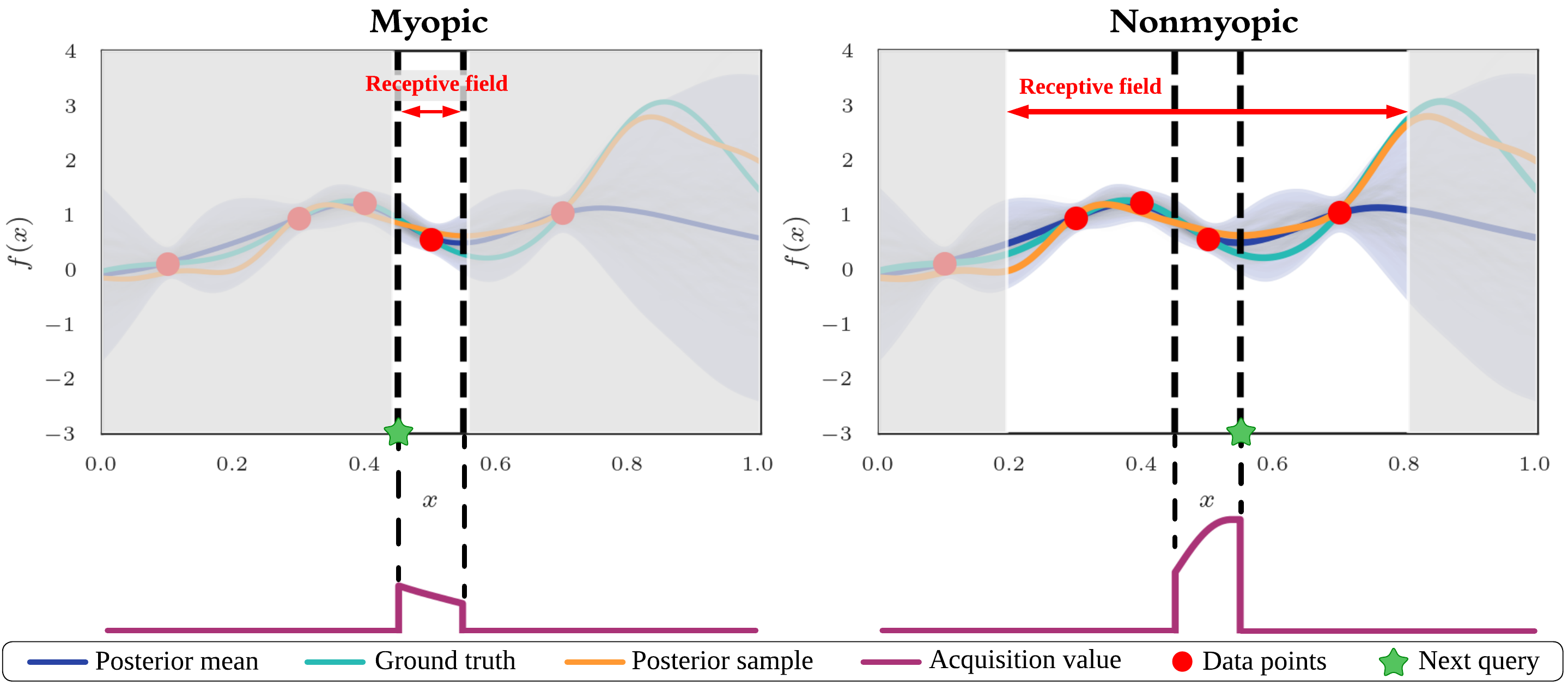}
    \caption{Comparison of myopic and nonmyopic BO in a dynamic cost setting. The unshaded region denotes the current receptive field, i.e., the subset of the input space considered for the next query. Myopic BO (left) has a narrow receptive field, focusing only on immediate reward. Nonmyopic BO (right) expands the receptive field by accounting for future queries, enabling selection of points that may appear suboptimal in the short term but lead to higher long-term rewards (see acquisition values in the bottom row). This broader planning horizon allows the decision-maker to access high-reward regions that greedy strategies cannot reach. See Section~\ref{sec:dynamic cost setting} for details.}
    \label{fig:main_fig}
\end{figure*}

\section{Introduction}
Bayesian optimization (BO) combines a probabilistic surrogate model, often a Gaussian process (GP), with an acquisition function to guide the search on black-box functions~\cite{shahriari2016boreview, frazier2018tutorial, garnett_bayesoptbook}. Standard BO typically assumes static query costs~\cite{1962_Kushner, 1964_Kushner}, but this assumption does not always practically hold. In many applications, these costs vary with prior actions. For example, in geological surveys, the cost of sampling a site depends on its distance from previous sites, reflecting transportation expenses~\cite{Bordas2020}. In biological sequence design, making a single token edit is inexpensive, whereas larger edits that move beyond a small edit distance are far more costly~\cite{Haiwei2004, Blazej2017}. Such environments introduce dynamic, history-dependent cost structures, where the cost of evaluating a point depends on the most recent query or even the full query trajectory~\cite{DCBO, 2021_Lee, folch2022snake, 2024_Folch}. Accounting for these costs fundamentally changes the optimization problem: the feasible decision space evolves with history, and effective strategies must plan over multiple steps rather than rely on purely myopic choices.

As cost structures become more complex and interdependent, myopic strategies fail to capture long-term benefits, highlighting the need for nonmyopic BO. Nonmyopic BO incorporates lookahead steps to improve current decision~\cite{2016_Gonzalez,budgetedBO,2020_Yue,2020_Jiang}. A natural way to formalize this problem is as a Markov Decision Process (MDP), which is commonly used in sequential decision-making~\cite{garcia2013markov, 2014_Puterman}. To determine the optimal next action in a sample-efficient manner, an MDP frames the decision process as a cost-constrained model-based reinforcement learning (CMBRL) problem, where the queried inputs are states and actions influence the transitions between consecutive states.

However, CMBRL exhibits two critical mismatches with nonmyopic BO. First, its reliance on simple neural network policies limits scalability to large, structured action spaces, as in biological sequence design, where edits carry semantic meaning and domain constraints~\cite{Janer2019,Wang2020Exploring,hafner2021mastering,Stolze2015,hamed2024dr}. Integrating pre-trained large language models (LLMs) as policies can exploit such structure, yet most existing reinforcement learning frameworks target single-step or contextual bandit regimes rather than multi-step nonmyopic BO~\cite{palo2023towards,zhuang2024toolchain,hazra2024saycanpay,Ouyang2024Training,trl,hu2024openrlhf,zheng2024llamafactory,HarperNemo}. Secondly, in domains requiring nonmyopic planning such as sequence editing, state transitions are often deterministic, reducing the need for stochastic world models~\cite{Janer2019,hafner2021mastering,Wang2020Exploring}. Moreover, CMBRL typically integrates neural reward models, which causes it to struggle to represent reward uncertainty, which is important for balancing exploration-exploitation~\cite{10.5555/3666122.3667947,10.5555/3666122.3669643,ZANGIROLAMI2024111663,Ez-zizi2023}. These reward models are often miscalibrated, yielding overconfident or underconfident estimates and, consequently, leading to suboptimal actions~\cite{minderer2021revisiting,10.5555/3540261.3541970,10.5555/3666122.3669643}. Explicitly modeling aleatoric and epistemic uncertainty via acquisition functions improves robustness to dynamic, noisy settings such as wet-lab biological testing, where small errors can noticeably change outcomes~\cite{10.1109/ICRA.2019.8793611,pmlr-v206-luis23a,Ez-zizi2023,10.1093/bib/bbv004}.

Recent advances in nonmyopic BO have achieved notable progress. While nonmyopic BO has been studied for decades, traditional methods can scale only up to a four-step lookahead~\cite{Snoek2012,2019_Wu,budgetedBO,oneshottree,2021_Lee} and typically ignore cost models. These approaches rely on free variables, leading to exponential complexity as the lookahead horizon increases. More recent work has sought to incorporate dynamic costs~\cite{ling_gaussian_2016,ijcai2023p446}, yet these methods still suffer from an exponential growth in optimization parameters. Neural policies, including transformer-based models, have shown promise by variationally optimizing the decision variables, thereby reducing the number of optimization parameters~\cite{maraval2023endtoend,nwankwo2024differentiating,cheon2025earlbo}. However, these approaches still require an exponential number of rollout trajectories during lookahead. A promising direction to alleviate this is decomposing actions into micro-actions, which approximates long-horizon lookahead with finer-grained steps~\cite{pmlr-v108-kharkovskii20a}. Nevertheless, splitting into macro-actions is less effective, as it inflates the action space, hinders correlation modeling, and reduces adaptivity without new observations.

To address these limitations,  we introduce LookaHES, a cost-constrained nonmyopic BO algorithm. LookaHES reduces the exponential complexity of multi-step optimization
by combining a multi-step variant of $H$-Entropy Search with pathwise sampling and neural policy optimization. Our approach balances exploration and exploitation using a Bayesian reward model and scales to horizons of twenty or more steps, well beyond the four-step limits of prior work. Additionally, LookaHES can be applied across diverse domains, from sequence design to natural language processing, where multiple interactions are required before a final decision is made. Our contributions are stated as follows.

\begin{itemize}
    \item We formulate nonmyopic BO in dynamic cost settings with cost models inspired by real-world scenarios such as protein sequence design.

    \item We employ a neural policy to variationally optimize all decision variables (i.e., reduce the number of optimization parameters) and pathwise sampling to reduce trajectory samples, enabling long-horizon planning and effective handling of large structured combinatorial action spaces.

    \item We evaluate LookaHES on nine synthetic benchmarks (2D–8D, varying noise) and a real-world geospatial optimization task using NASA night-light imagery, showing consistent improvements over myopic and nonmyopic baselines.
    
    \item We demonstrate the effectiveness of LookaHES on a constrained protein sequence design task, where combinatorial complexity hampers traditional BO. Leveraging a LLaMa-3.2-based policy, our method exploits pretrained domain knowledge to guide edits, achieving superior performance.
\end{itemize}

%% file: sections/3.cost_taxonomy.tex
\section{Related Works}
\label{sec:cost_taxonomy}
We review related work on cost-sensitive BO by introducing a taxonomy that classifies costs into four categories based on their uncertainty (known vs. unknown) and variability (static vs. dynamic), as summarized in Table~\ref{table:cost_types}. Within static costs, prior work distinguishes between homogeneous costs, where all queries incur the same expense, and heterogeneous costs, where the cost depends on the specific query. To illustrate the distinction between cost structures, we visualize the uncertainty and variability of these structures as probabilistic graphical diagrams in Figure~\ref{fig:cost_diagram}. Additional related works are provided in Appendix~\ref{further_related_work}.

\begin{table*}[!htbp] 
    \centering 
    \caption{Classification of Cost Types Based on Uncertainty and Variability.} 
    \label{table:cost_types} 
    \begin{tabular}{b{0.25cm}|p{7.75cm}|p{8.15cm}}
        \textbf{} & \textbf{Known Cost} & \textbf{Unknown Cost} \\ 
        \midrule
        \multirow{3}{*}{\rotatebox{90}{\textbf{Static}}} & 
        Costs remain constant regardless of the queries made during optimization. These costs are predictable and can be pre-determined, making them straightforward to budget and plan~\cite{2019_Wu, nyikosa2018bayesianoptimizationdynamicproblems, 2016_Lam}. &
        Costs are static but unpredictable due to external factors such as system fluctuations or resource availability. While they are static, their unpredictability complicates cost estimation~\cite{budgetedBO, 2021_Lee, Snoek2012, LUONG2021107481}. \\ 
        
        \midrule
        \multirow{3}{*}{\rotatebox{90}{\textbf{Dynamic}}} & 
        Costs change based on the sequence of previous queries. These costs depend on previous optimization steps but remain predictable, allowing for some level of planning~\cite{ijcai2023p446,ling_gaussian_2016}. &
        Costs depend on prior queries and are unpredictable, often arising in variable environments like dynamic resource allocation or uncertain execution times, which are difficult to estimate in advance. To our knowledge, no prior work addresses this.
    \end{tabular}
\end{table*}

Prior work has considered unknown heterogeneous costs, particularly in hyperparameter optimization, where costs are static but configuration-dependent~\cite{budgetedBO,2021_Lee,LUONG2021107481,Snoek2012}. Other approaches assume constant, known costs across the search space~\cite{2019_Wu,nyikosa2018bayesianoptimizationdynamicproblems,2016_Lam}, thereby ignoring cost variability. In contrast, our study addresses Bayesian optimization with known, dynamic costs, a regime not much explored in earlier work. For instance,~\cite{ijcai2023p446,ling_gaussian_2016} introduce Euclidean cost-constrained lookahead acquisition functions related to our setting. However, these methods fail to overcome the exponential growth in decision variables and rollout trajectories, limiting their scalability to long horizons. Our proposed LookaHES mitigates these limitations by employing a language-model-based policy to reduce the number of optimization parameters and pathwise sampling to control the number of rollout trajectories.

%% file: sections/4.method.tex
\section{Method}
\label{sec:method}

\subsection{Problem Setup}
The decision maker (DM) aims to optimize a black-box function $f^: \Xc \rightarrow \Yc$, with $\Yc \subseteq \Rbb$, by sequentially issuing $T$ queries. At each step $t \in \{1,\dots,T\}$, the DM selects a query $x_t \in \Xc$ and observes a noisy evaluation $y_t = f^(x_t) + \epsilon_t$. The complete history of queries and observations is denoted by $x_{1:T} = [x_1, \dots, x_T]$ and $y_{1:T} = [y_1, \dots, y_T]$, respectively. Given a prior distribution over the parameters $p(\theta)$, a surrogate model $f_{\theta}$ of the black-box function $f^*$ is sampled by $\theta \sim p(\theta)$. At time step $t$, given the history of queries and observations, the function parameters' posterior distribution is $p_t(\theta) = p(\theta | D_t) = p(\theta | x_1, y_1, \dots, x_t, y_t)$.

After $T$ queries, DM selects an action $a \in \Ac$. In general Bayesian decision-making literature, $\Ac$ can be distinct from $\Xc$. For example, the DM might query the black box function to find the top-$k$ or level set. This paper focuses on global optimization; hence, the action set is the same as the query set, $\Ac = \Xc$\footnote{We still use the notion of a final action to maintain consistency with the literature. We note that our method can apply to a general action set, and studying that is beyond the scope of our paper.}. The loss function is given by $\ell(f^*, a) = -f^*(a)$. Following prior work~\cite{russo2016information, pmlr-v84-kandasamy18a}, the Bayesian cumulative regret at timestep $T$ is defined as $\Ebb \left[ \sum_{t=1}^{T} \big(f^*(a^*) - f^*(a_t)\big) \right]$, where the expectation is taken over the randomness from the environment, the sequence of queries, and the final actions. We summarize our notations in Appendix~\ref{sec:notation}.

\subsection{Decision-making Objective}
The DM selects queries via an acquisition function. We consider $H$-Entropy Search (HES)~\cite{degroot1962uncertainty, neiswanger2022generalizing}, a decision-theoretic entropy-based framework that generalizes common acquisition functions such as Knowledge Gradient~\cite{2009_Frazier} and Expected Improvement~\cite{Saltenis1971}. We focus on its multi-step lookahead variant, which accommodates dynamic cost structures $c$, and describe our procedure for optimizing a variational version of the acquisition function using a neural network policy.

For a prior $p(f)$ and a dataset $D_t = D_0 \cup \{ (x_i, y_i )\}_{i=1}^{t}$, the posterior $\Hbb_{\ell, c, \Ac}$-entropy and the expected $\Hbb_{\ell, c, \Ac}$-information gain (EHIG) at step $t$ with loss function $\ell$, cost function $c$, action set $\Ac$, and Lagrange multiplier $\lambda$, and lookahead horizon $L$ is
\begin{equation*}
    \begin{aligned}
    \Hbb_{\ell, c, \Ac} [f | D_t] &= \inf_{a \in \Ac} \{ \Ebb_{p_t(f)} [\ell(f, a)] + \lambda c(x_{1:t}, a) \}, \\
    \text{EHIG}_t (x_{1:L}) &= \Hbb_{\ell, c, \Ac} [f | D_{t}] \\
    &- \Ebb_{p_t(y_{1:L}|x_{1:L})} \left[\Hbb_{\ell, c, \Ac} [f | D_{t+L}] \right].
    \end{aligned}
\end{equation*}
Following the $\Hbb$-information gain heuristics, the decision-maker selects the query $x_{t+1} \in \Xc$ at each step $t$ to maximize the expected $\Hbb$-information gain:
\begin{equation}
\label{acq_function}
    \begin{aligned}
    x_{1:L}^* &= \argsup_{x_{1:L} \in \Xc^L} \text{EHIG}_t (x_{1:L})\\
    &= \argsup_{x_{1:L} \in \Xc^L} \left[ - \Ebb_{p_t(y_{1:L}| x_{1:L})} [\Hbb_{\ell, c, \Ac} [f | D_{t+L}]] \right].\\
    \end{aligned}
\end{equation}

\subsection{Dynamic Cost Bayesian Optimization}
\label{sec:dynamic cost setting}
\begin{figure*}[htb!]
    \centering
    \includegraphics[width=0.7\linewidth]{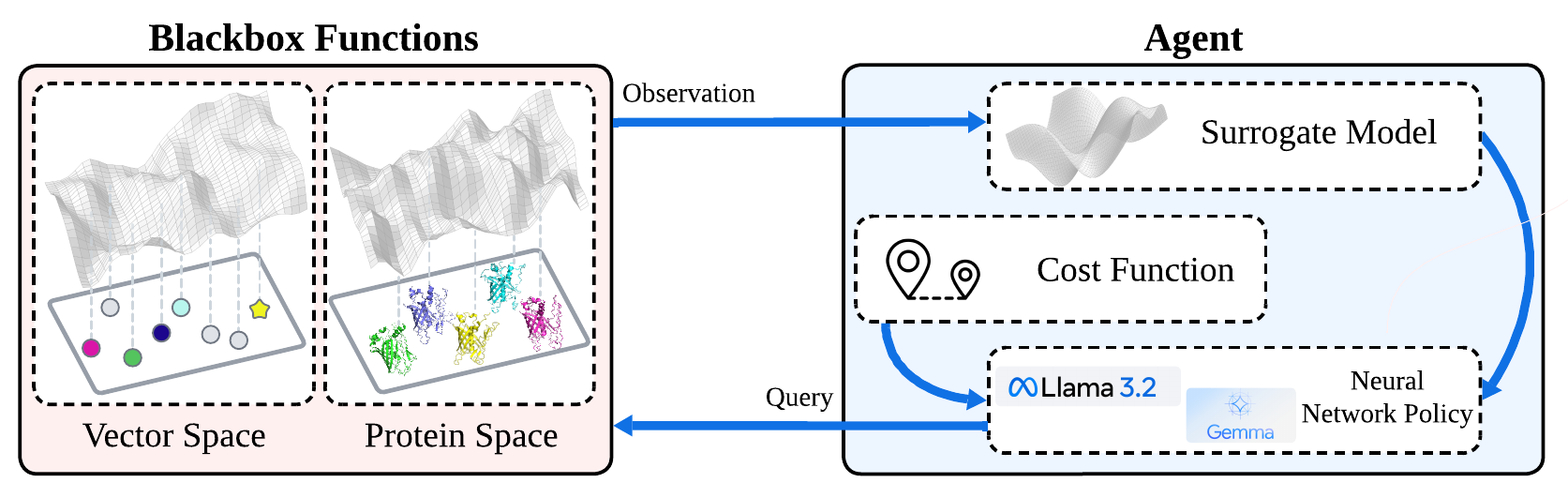}
    \caption{Process illustration. The black-box functions define complex objective landscapes in vector or protein spaces. The agent leverages a surrogate model to approximate the black-box function. The neural network policy generates the next queries to explore and exploit the optimization space. The query generation is guided by a dynamic cost function, ensuring efficient and targeted navigation of the search landscape.}
    \label{fig:overviewBO}
\end{figure*}

To model the dynamic cost, firstly, we define the cost of querying $x_t$ as $c(x_{<t}, x_t)$, where $c$ is an application-specific cost function provided to the decision-maker. The total cost to execute $T$ queries is $\sum_{t=1}^T c(x_{<t}, x_t)$. We introduce two primary dynamic cost structures: \textbf{(i) Markovian cost}, which depends only on the previous query $x_{t-1}$ and the current query $x_t$, typically based on the $p$-norm distance between them. Costs may be nonlinear, e.g., traveling within a radius $r$ may be free, but beyond that, it grows at a rate $k$, and observed costs may include noise $\epsilon$: $c_{\text{Markov}}(x_{t-1}, x_t) = \max(k(||x_t - x_{t-1}||_p - r), 0) + \epsilon$. Common instances include Euclidean cost ($p=2, r=0$) as in ground surveys~\cite{Bordas2020}, Manhattan cost ($p=1, r=0$), and spotlight cost ($k=\infty$), e.g., in biological sequence design where more than one token cannot be edited in a single experiment~\cite{belanger2019biological}. \textbf{(ii) Non-Markovian cost}, which depends on the full query history, e.g., a traveler may receive a discount $d$ if cumulative distance exceeds $m$, formalized as $c_{\text{non-Markov}}(x_{<t}, x_t) = c_{\text{Markov}}(x_{t-1}, x_t) - d \Ibb[\sum_{i=1}^{t-1} c_{\text{Markov}}(x_i, x_{i+1}) > m]$. In practice, cost models can be learned from application data or designed by the decision maker. Under budget constraints, dynamic costs require nonmyopic planning to avoid high costs or local optima.

Importantly, under dynamic costs, \textbf{candidates in the query space $\Xc$ are not equally considered} by the DM. For instance, in the spotlight cost, candidates outside the spotlight radius are effectively inaccessible. Looking ahead allows the DM to plan over future steps, considering candidates that may \textit{become accessible after intermediate queries}. We define the decision receptive field as the subset of the input space considered by the DM. By looking ahead, the receptive field expands, allowing the DM to ``invest'' in suboptimal immediate queries if they enable better outcomes later, such as access to the global optimum (Figure~\ref{fig:main_fig}). The longer the lookahead horizon is, the better the DMs plan their decision. Unfortunately, increasing the lookahead horizon leads to an exponential growth in decision variables and uncertainty, making planning intractable.

\subsection{Neural Network Policy Optimization}
In nonmyopic decision-making, the number of decision variables scales with the number of Monte Carlo (MC) rollout paths, $p$, and the horizon length, $T$. In the best-case scenario, where the number of MC samples grows linearly with the horizon, the policy complexity is $\Oc(T)$. In the worst case, the number of samples grows exponentially, leading to complexity $\Oc(k^T)$, where $k$ is the number of samples per step. To mitigate this issue, we employ a neural network policy as it can reduce the number of optimization parameters to a constant with respect to the horizon. These policies have been widely applied in contexts such as policy gradient methods, variational autoencoders, and variational design of experiments~\cite{schulman2017proximal,kingma2022autoencodingvariationalbayes,Foster2019Variational}.

From \eqref{acq_function}, we observe that for each lookahead step $l \in [L]$, the decision $x_{t+l+1}$ is determined by the previous decision variables and corresponding observations, $(x_{1:t+l}, y_{1:t+l})$. This dependency can be modeled using a recurrent neural network (RNN) parameterized by $\xi \in \Xi$, which takes the history as input to predict the optimal next query: $\xi: (x_{1:t}, y_{1:t}) \mapsto x_{t+1}$. The corresponding posterior predictive $y_{t+1}$ can then be computed by $y_{t+1} \sim p_{t}(x_{t+1})$. We maintain the gradients $\frac{\partial \text{EHIG}_t(x_{1:L})}{\partial \xi}$ for optimizing $x^*_{t+1} = \xi^*(x_{1:t}, y_{1:t})$ across the lookahead steps by applying the chain rule:
\begin{equation*}
     \frac{\partial \text{EHIG}_t(x_{1:L})}{\partial x_{t+L}} \frac{\partial x_{t+L}}{\partial \xi} + \frac{\partial \text{EHIG}_t(x_{1:L})}{\partial y_{t+L}} \frac{\partial y_{t+L}}{\partial x_{t+L}} \frac{\partial x_{t+L}}{\partial \xi}.
\end{equation*}
We can rewrite equation~\eqref{acq_function} as:
\begin{equation}
\label{equ:opt_obj}
    \begin{aligned}
    \xi^* &= \arginf_{\xi \in \Xi} \Ebb_{p_t(y_{1:L}| x_{1:L}, \xi)} \big[ \inf_{a \in \Ac} \ \Ebb_{p_{t+L}(f)} [\ell(f, a)] \\
    &\hspace{12em} + \lambda c(x_{1:t}, x_{1:L}, a) \big].
    \end{aligned}
\end{equation}
In our experiments, the variational network is trained using fantasized data points. Specifically, when optimizing step $t+1$, we use the previously observed data points $(x_{1:t}, y_{1:t})$ to generate lookahead data points $(x_{t+1:t+L}, y_{t+1:t+L})$ through an autoregressive process: $x_{t+l+1} = \xi_t(x_{1:t+l}, y_{1:t+l})$ and $y_{t+l+1} \sim p_{t+l}(x_{1:t+l})$. These fantasized data points are then used to compute the optimization objective and find the optimal $\xi_{t+1}$.

Predicting the next query is autoregressive, which makes language models suitable policies. Model architectures range from LSTMs~\cite{hochreiter1997long} to transformers such as BERT~\cite{devlin-etal-2019-bert} and decoder-only LLMs (e.g., GPT-4, LLaMA 3). As neural networks, these models can be designed to handle inputs and outputs in either continuous or discrete spaces, making them applicable to various domains. This allows them to support both continuous and discrete $\Xc$ and $\Ac$. In discrete domains such as drug design, where molecules are represented as strings, LookaHES embeds tokens into continuous space (analogous to sentence embeddings in NLP). Differentiation through discrete outputs is handled using the reparameterization trick~\cite{kingma2022autoencodingvariationalbayes} for small action sets, or policy-gradient methods such as REINFORCE~\cite{Williams1992}, which employ the log-derivative trick, for larger and more complex ones~\cite{JMLR:v21:19-346}.

\subsection{Pathwise Sampling for Trajectory Rollout}

We use pathwise sampling for the posterior predictive distribution to reduce the number of trajectories from exponential to constant. Specifically, we factorize $p(y_{1:T} \mid x_{1:T}, D_0) = \int p(f \mid D_0) \prod_{t=1}^T p(y_t \mid x_t, f) \, \mathrm{d}f$. At each step $t$, we draw a sample $f_t \sim p(f \mid D_t)$ and then generate $y_t = f_t(x_t)$. To approximate the posterior predictive, we begin with a fixed number of restarts $r$, so the number of trajectories equals the number of restarts rather than growing exponentially with $T$ (see Figure~\ref{fig:mc_tree}). We discuss the complexity trade-offs in Appendix~\ref{sec:pathwise_sampling}. Objective~\eqref{equ:opt_obj} can be formulated as
\begin{equation*}
\xi^* = \arginf_{\xi \in \Xi} \frac{1}{r} \sum_{\tau=1}^r 
    \Bigg[ \inf_{a \in \Ac} \big[ \ell(f_{t}^{(\tau)}, a)
    + \lambda c(x_{1:t}, x_{1:L}, a)  \big] \Bigg],
\end{equation*}
where $f_t^{(\tau)} \sim p(f \mid D_t)$, and $y_{1:L}^{(\tau)} = f_t^{(i)}(x_{1:L})$.

%% file: sections/5.experiment.tex
\begin{figure*}[htb!]
    \centering
    \includegraphics[width=\linewidth]{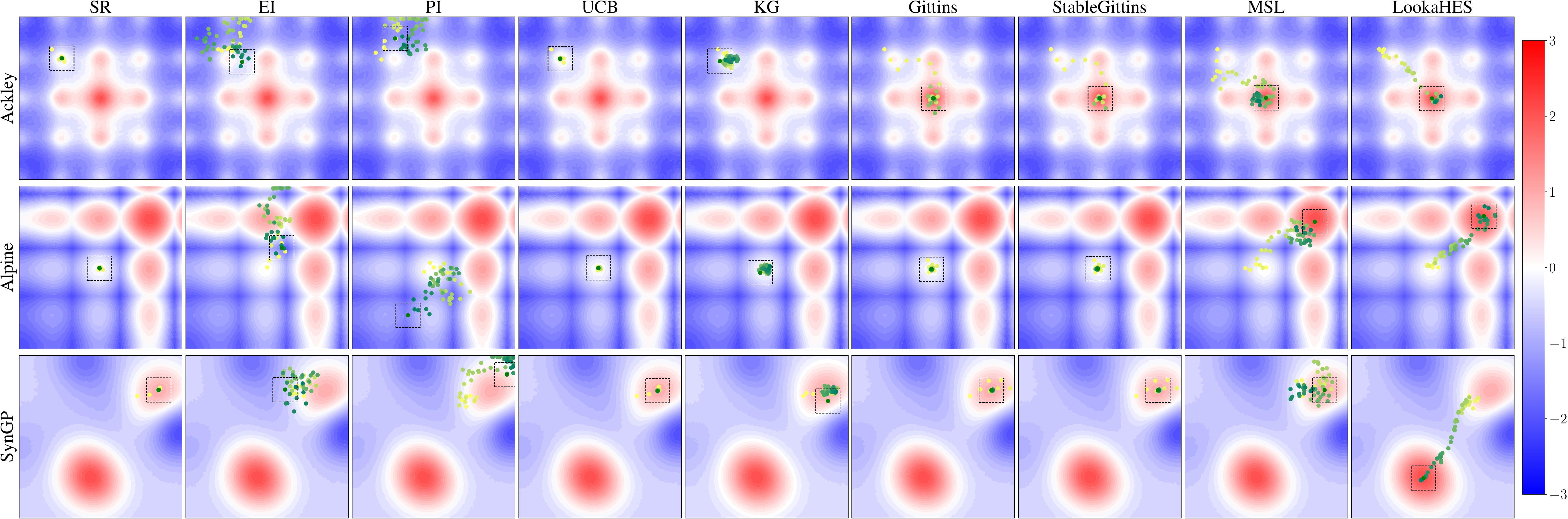}
    \caption{Queries across BO iterations with $\sigma=0.05$ and $r$-spotlight cost. Yellow and green points indicate the initial position and final action, respectively. LookaHES reaches the global optimum, whereas the others tend to be trapped in local optima.}
    \label{fig:query}
\end{figure*}
\section{Experiments and Results}
In the following section, we compare the performance and robustness of LookaHES through a series of experiments designed to address three objectives. First, we demonstrate the algorithm’s ability to efficiently optimize in environments with dynamically varying query costs. Second, we benchmark its performance against established baselines across both synthetic and real-world scenarios. Third, we highlight its practical advantages in terms of solution quality and computational efficiency. Collectively, these experiments aim to answer the following research questions.

\begin{itemize}
    \item \textbf{RQ1:} How does the proposed method compare to state-of-the-art myopic and nonmyopic acquisition functions in the continuous input domain under dynamic cost constraints?
    \item \textbf{RQ2:} Is it possible to apply the proposed method to problems with discrete input spaces?
    \item \textbf{RQ3:} How do aleatoric and epistemic noises, the quality of the surrogate model, and the lookahead horizon impact the performance of LookaHES?
    \item \textbf{RQ4:} Does `optimism' in myopic methods lead to better performance than nonmyopic methods without optimism? Can this optimism be broadly applied to real-world problems?
\end{itemize}

We compare LookaHES with eight baselines implemented with BoTorch~\cite{botorch} including Simple Regret (SR)~\cite{Zhao2023Revisiting}, Expected Improvement (EI)~\cite{Mockus1989BayesianAT}, Probability of Improvement (PI)~\cite{1964_Kushner}, Upper Confidence Bound (UCB)~\cite{2010_Srinivas}, Knowledge Gradient (KG)~\cite{2009_Frazier}, Gittins, StableGittins~\cite{Qian2024Cost}, and Multistep Tree (MSL)~\cite{oneshottree}. Details of these baselines are presented in Appendix~\ref{sec:detail_baseline}. All acquisition function values are estimated via the quasi-Monte Carlo method with the Sobol sequence~\cite{botorch}. We experiment with four cost functions, including Euclidean, Manhattan, $r$-spotlight, and non-Markovian cost based on Euclidean distance. We use the Sample Average Approximation with a base sample as a variance reduction technique, which significantly improves the stability of the optimization. All optimizations are conducted with 64 restarts, employing the default value of $\lambda = 1$ to regulate the DM's budget allocation. The lookahead horizon is set to 20 for both LookaHES and Multistep Tree. Each experiment is repeated with three random seeds. All experiments are performed on an NVIDIA A100 GPU 80GB.

\subsection{Continuous Synthetic Functions}
To answer \textbf{RQ1}, we evaluate LookaHES on nine synthetic functions for global optimization in continuous spaces. The 2-dimensional functions, with their initial data points and maximum BO steps, include Ackley (50 samples, 100 steps), Alpine (100 samples, 50 steps), HolderTable (100 samples, 50 steps), Levy (100 samples, 50 steps), Styblinski-Tang (50 samples, 50 steps), and SynGP (25 samples, 50 steps). The SynGP function is sampled from a 2D RBF Gaussian Process with length scale $\sqrt{0.25}$ and signal variance $1$. High-dimensional functions include Ackley4D (4D, 100 samples, 100 steps), Hartmann (6D, 500 samples, 100 steps), and Cosine8 (8D, 200 samples, 100 steps). More details of these functions are available in~\cite{BinghamOptimization}. The variation in the number of initial samples and optimization steps reflects the complexity of each function. To evaluate the robustness of our method, we conduct ablation studies on noise levels, initial data points, surrogate model kernels, lookahead horizon, and hyperparameter choices in the myopic acquisition function, as detailed in Appendix~\ref{sec:ablation_synthetic}.

\begin{figure}[htbp]
    \centering
    \includegraphics[width=\linewidth]{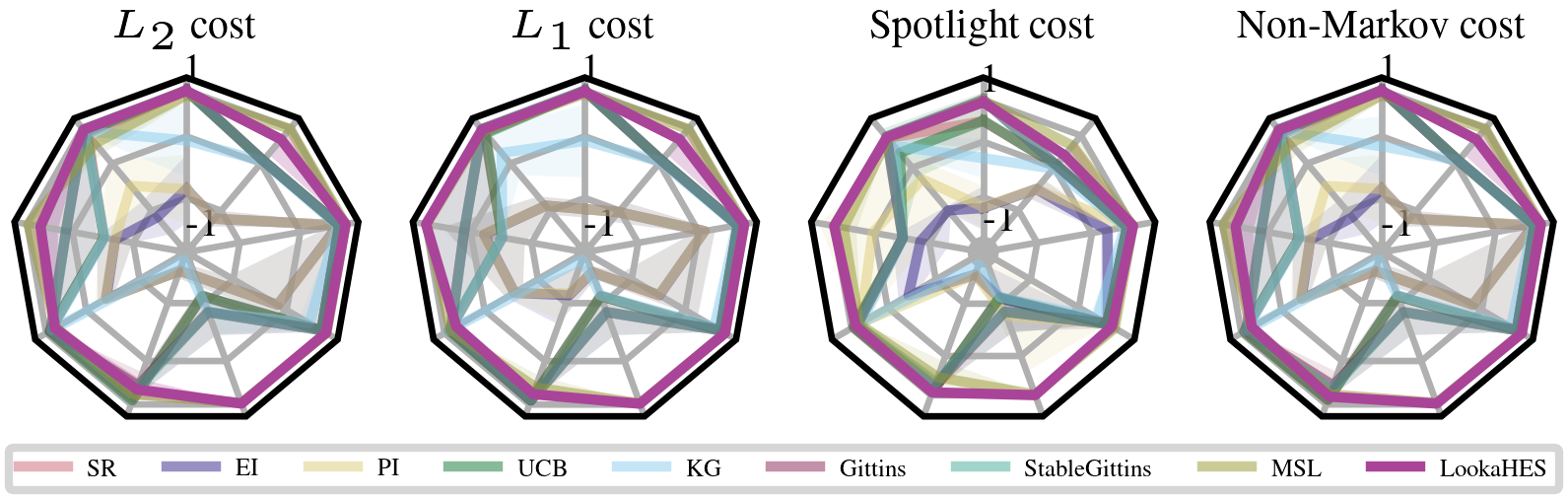}
    \caption{Final observed value at $\sigma=0.05$. From the north of each plot, counter-clockwise: Ackley, Ackley4D, Alpine, Cosine8, Hartmann, HolderTable, Levy, StyblinskiTang, SynGP. LookaHES consistently found global optimum across various cost structures.}
    \label{final-value}
\end{figure}

All function inputs are normalized to the hypercube $[0,1]^d$, and outputs are scaled to the range $[-3, 3]$. The global maximum of each function is 3, so the instantaneous regret of an action $a$ is defined as $3 - f^*(a)$. Observations are corrupted with three noise levels: 0\%, 1\%, and 5\%. We use a Gaussian Process with a Matern kernel as the surrogate model. The variational neural network consists of a two-layer encoder, a Gated Recurrent Unit, and a three-layer decoder with exponential linear unit activations and 64 hidden dimensions. This network is trained using Adam optimizer with $10^{-3}$ learning rate. During inference, we add a small noise sampled from the von Mises–Fisher distribution~\cite{Fisher1953} to the predicted query to enhance the exploration and facilitate acquisition function optimization restart. Figure~\ref{fig:query} illustrates that myopic algorithms often converge to local maxima due to their inability to account for long-term effects, leading to suboptimal solutions that seem beneficial in the short term. In contrast, LookaHES can consider future outcomes, enabling it to move toward the global maximum. With our improvements, the MSL method can achieve a lookahead of up to 20 steps, yielding outcomes similar to LookaHES. However, without the neural policy, MSL optimizes directly on decision variables, which limits its applicability to real-world scenarios, as demonstrated in the next section. Figure~\ref{final-value} compares the baseline methods and our proposed approach in terms of the final observed values under the highest noise level (\(\sigma = 0.05\)) across four cost functions. We also evaluate LookaHES to a real-world problem in continuous space for identifying areas with the most light in NASA night-light imagery. The results of this experiment are presented in Appendix~\ref{sec:nightlight}.

\begin{mybox}
\textbf{\textit{Summary:}} LookaHES consistently outperforms the myopic and is comparable to nonmyopic baselines on synthetic functions, highlighting the advantage of a long lookahead horizon.
\end{mybox}

\subsection{Protein Sequence Design}
\begin{figure}[!htbp]
    \centering
    \includegraphics[width=0.8\linewidth]{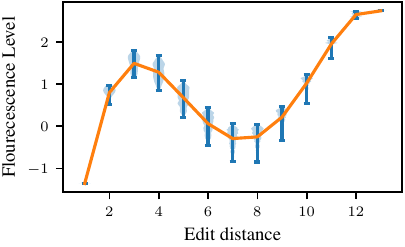}
    \includegraphics[width=\linewidth]{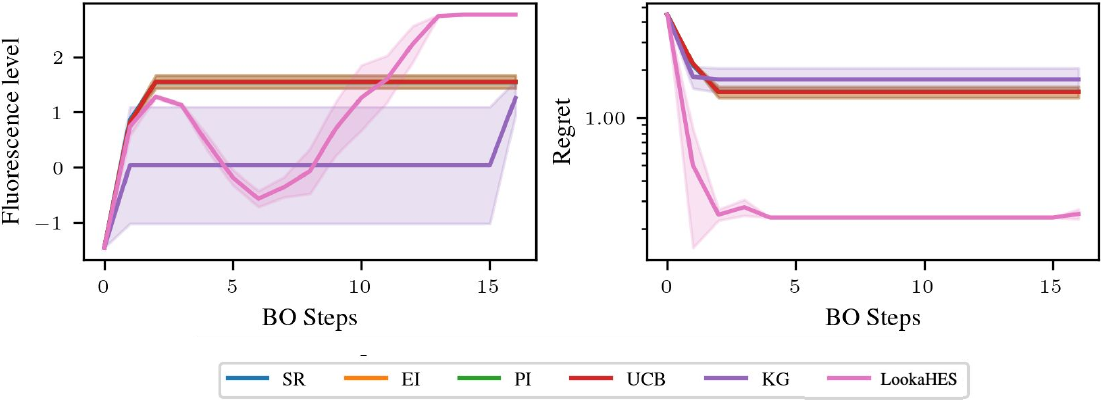}
    \caption{Fluorescence distribution by edit distance (top), observed fluorescence across BO steps (bottom left), and regret across BO steps (bottom right). Myopic methods are trapped in local minima ($\approx 1.5$ fluorescence), while our 12-step LookaHES anticipates the global maximum, achieving $\approx 2.7$ fluorescence.}
    \label{fig:yX_by_step}
\end{figure}

We demonstrate the practical applicability of LookaHES to optimize protein sequences~\cite{elnaggar2023ankh}. The decision-making process involves determining whether to edit a given protein sequence at a single position. The spotlight cost function is applied in this experiment to simulate real situations where editing more than one position will incur unpredictable, large costs. This experiment addresses \textbf{RQ2} with real-world problems. Figure~\ref{fig:yX_by_step} (top) provides a visualization of the protein space. If editing is chosen, the next step is to determine the position to be edited and select the new amino acid. We conduct a sequence of $T = 12$ edits to maximize the fluorescence level obtained from a wet lab experiment, given by the black-box oracle $f^*: \Xc \rightarrow \Rbb$, which is expensive to query. We assume that $f^*$ has a parametric functional form on the feature space $\phi(x)$: $y = f_{\theta^*}(x) = g(x) + \alpha(\phi(x)^\top\theta^* + \epsilon)$, where $\theta^* \sim p(\theta) = \Nc(\mu, \Sigma)$, $\epsilon \sim \Nc(0, \sigma)$, $g(\cdot)$ is a synthetic function, and $\alpha$ is a scaling hyperparameter. %

We utilize the ProteinEA Fluorescence dataset~\cite{proteinea_fluorescence}, which comprises 21,445 training samples, to construct the black-box oracle. We experiment with various feature extraction functions $\phi(\cdot)$, including LLaMa-2 7B~\cite{touvron2023llama}, LLaMa-3 8B~\cite{llama3modelcard}, Mistral 7B~\cite{jiang2023mistral}, Gemma 7B~\cite{gemmateam2024gemma}, ESM-2 650M, ESM-2 3B~\cite{lin2022language}, and Llama-Molist-Protein 7B~\cite{fang2024molinstructions}. Figure~\ref{fig:goodness_of_fit} shows validation results of the parametric black-box oracle with varying training sample sizes. Gemma 7B achieves the highest validation $R^2$ for predicting fluorescence, so we use it as the feature extraction for all methods.

Next, we construct our semi-synthetic protein space using a sequence from the ProteinEA Fluorescence. Specifically, we select a single sequence from the validation set consisting of 237 amino acids across 20 types. In this experiment, the protein designer can edit only one amino acid at a time across a maximum of 12 fixed positions and is limited to 2 possible amino acid types for each position. Under this setting, the protein space $\Xc$ contains $|\Xc| = 4096$ possible proteins. We then compute the fluorescence values for these proteins using the previously constructed oracle. The goal is to edit a starting protein to achieve the highest fluorescence, defined as $x_{max} = \argmax_{x_i \in \Xc} \Ebb_{\theta^* \sim p(\theta)} f_{\theta^*}(x_i)$. The starting protein, $x_0$, is chosen as the one with an edit distance of 12 from the protein with maximal fluorescence. Because each edit position can only accommodate two different tokens in this setup, there is only one possible starting protein. We choose $\alpha = 0.2$ and $g(x) = -0.005(d - 0.5)(d - 5)(d - 8)(d - 13.4)$, where $d = d_{edit}(x, x_0)$ represents the edit distance between the starting protein and a given protein $x$. We ablate this experiment with a different starting protein and synthetic function $g(x)$. The ablation results are presented in Appendix~\ref{sec:protein_exp_ablation}.

We use a Bayesian linear regression model as a surrogate to guide optimization. At each step $t \in [T]$, hyperparameters are estimated via maximum marginal log-likelihood, and the next mutation is selected by optimizing the language model (Section~\ref{sec:method}). This process maximizes fluorescence, maintaining the cost constraints. The mutation and fluorescence value are given by $x_{t+1} \sim p_{\xi_t}(x_t)$ and $f_{\theta^*}(x_{t+1})$, respectively, updating the dataset before the next iteration. This optimization can be batched, and we use Llama-3.2 3B as the variational network. We set the online and lookahead steps to $16$ and $12$, respectively. The experiments are repeated with three seeds. Before starting the optimization process, the model is supervised finetuned with random mutation data. In online iteration, we optimize the network with a maximum of 768 gradient steps and perform 64 restarts to select the next mutation. To enhance optimization efficiency, we modified Proximal Policy Optimization (PPO) by decoupling training and inference, with vLLM~\cite{kwon2023efficient} handling the lookahead rollout. After each gradient update, network weights are transferred to vLLM for the next rollout. To accommodate the spotlight cost during the generation of each lookahead sequence, we employ an iterative approach, attempting to regenerate the sequence up to 32 times. If regeneration is unsuccessful, we randomly mutate to the most recent sequence, with a 50\% probability of retaining it unmodified. Figure~\ref{fig:yX_by_step} (bottom left) shows the observed fluorescence levels, while the (bottom right) side presents the regret ($3 - f_{\theta^*}(a)$) across online iterations. More details about our experiments can be found in Appendix~\ref{sec:protein}.

Under the single-edit constraint, myopic acquisitions stall in a local basin ($\approx 1.5$ fluorescence), whereas our 12-step LookaHES leverages bridge edits to reach the $\approx 2.7$ mode with lower cumulative regret. Mechanistically, lookahead enlarges the decision receptive field, leading the policy to prioritize downstream access over one-step gains. As shown in Figure~\ref{fig:yX_by_step}, LookaHES advances across edit-distance steps before descending into the global basin, whereas myopic methods oscillate locally and repeatedly sample low-yield variants, missing the nonlocal route to the optimum.

\begin{mybox}
\textbf{\textit{Summary:}} LookaHES works well on discrete domains, including the protein editing task, with superior performance to other myopic approaches, proving its effectiveness across diverse applications.
\end{mybox}

%% file: sections/6.discussion.tex
\section{Discussion, Limitations, and Future Work}
We propose LookaHES for the nonmyopic BO in dynamic cost settings. LookaHES incorporates dynamic costs and downstream utility, leading to more informed decision-making under uncertainty. By utilizing a neural network policy, it achieves scalability in planning multiple steps ahead. Experimental results demonstrate its outstanding performance compared to baseline methods on various benchmarks. However, the method has limitations. It requires a well-defined cost model upfront, which may not always be practical, and its performance relies heavily on a well-specified surrogate model. Future work should explore the impact of model misspecification on plan quality. Further advancements in adaptive modeling and cost estimation could enhance its robustness, expanding its applicability to even more complex decision-making scenarios.

%% file: sections/8.appendices.tex
\section{Notation}
\label{sec:notation}

Table~\ref{tab:notation} is a glossary of the mathematical notation used in the paper.
\begin{table}[!htbp]
    \centering
    \caption{Glossary of Mathematical Notation}
    \label{tab:notation}
    \begin{tabular}{>{\raggedright\arraybackslash}p{1.5cm}>{\raggedright\arraybackslash}p{10cm}}
        \toprule
        \textbf{Symbol} & \textbf{Description}\\
        \midrule
        $f^*$ & Black-box function \\
        $p(f)$ & Prior distribution over the black-box function \\
        $\theta$ & Random variable representing the parameters of the black-box function in the parametric form \\
        $\theta^*$ & Optimal parameters of the black-box function in the parametric form \\
        $\Xc$ & Input domain \\
        $\Yc$ & Output domain \\
        $\xi$ & Parameters of the variational network \\
        $D_t$ & $D_t = \{(x_i, y_i)\}^t_{i=1}$ Dataset acquired \\
        $p_t(\cdot)$ & The posterior distribution conditioned on the data up to and including timestep $t$ \\
        $c(\cdot, ..., \cdot)$ & $c: \Xc^k \rightarrow \Rbb$, cost function depending on $k$-step history of query \\
        $[T]$ & $\{1, ..., T\}$ \\
        $\Hbb_{\ell, \Ac}$ & The decision-theoretic entropy~\cite{degroot1962uncertainty} corresponding to a loss function $\ell$ and an action set $\Ac$ \\
        $L$ & Lookahead horizon \\
        $T$ & Number of interactions with the environment \\
        \bottomrule
    \end{tabular}
\end{table}

\begin{figure}[!hbt]
    \centering
    \includegraphics[width=0.7\linewidth]{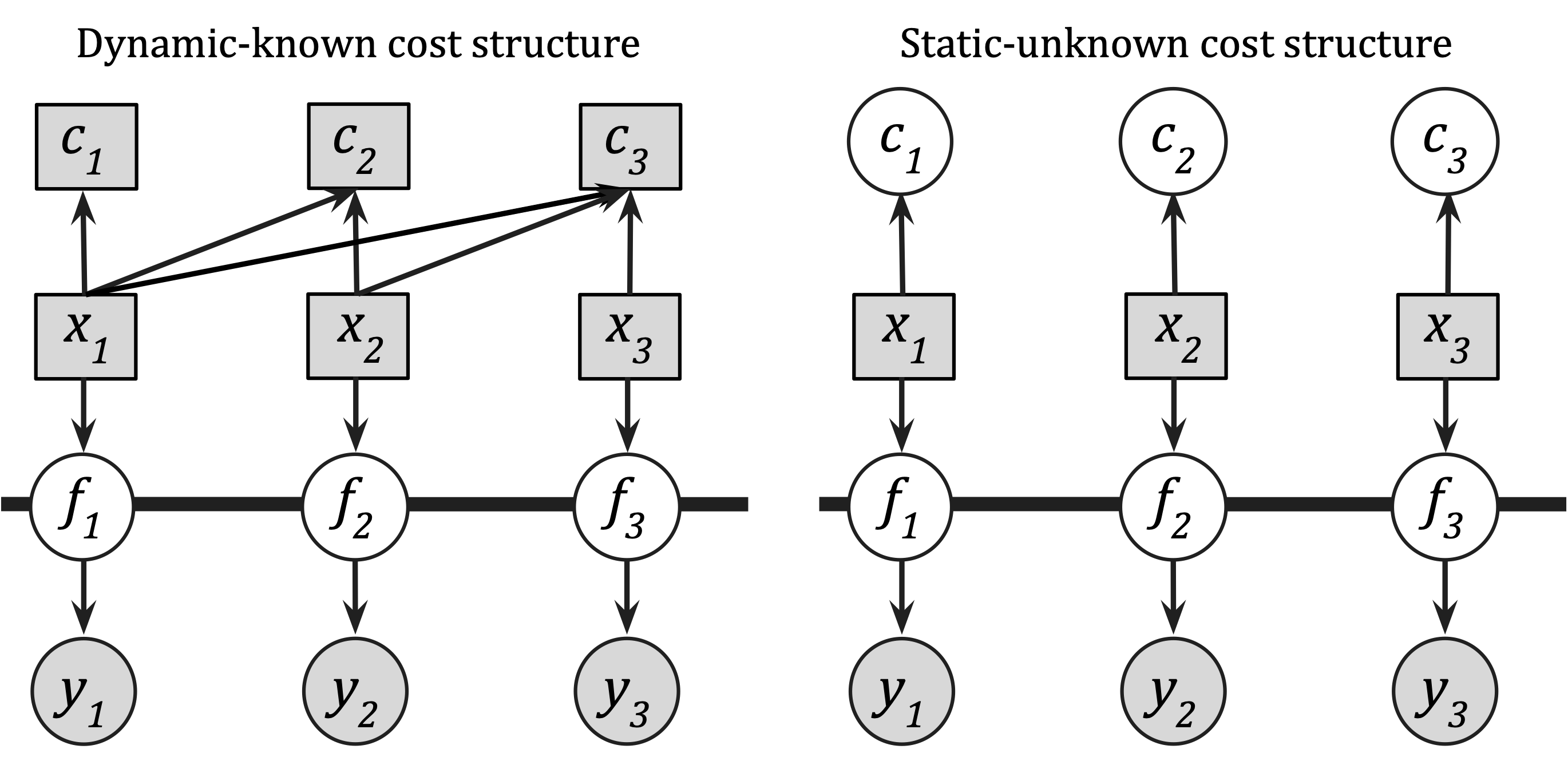}
    \caption{Graphical models of popular cost structures. In this diagram, $f$ represents the target black-box function, $x$ denotes the input query, $y$ is the output value, and $c$ is the cost of querying $x$. On the left — the dynamic-cost structure — the cost of querying $x_3$ can depend on $x_1$ and $x_2$. On the right — the static-known cost structure — the cost of querying $x_3$ is independent of other queries.}
    \label{fig:cost_diagram}
\end{figure}

\input{sections/3.related_work}

\section{Pathwise Sampling}
\label{sec:pathwise_sampling}
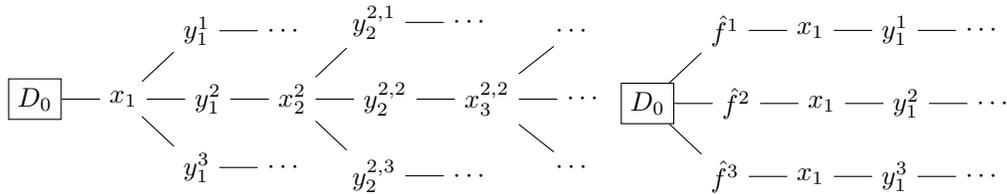
\begin{figure}[htbp]
\centering
\resizebox{.8\textwidth}{!}{
    \begin{tikzpicture}[node distance=0.5cm]
        \node[draw, rectangle] (D) {$D_0$};
        \node[right=of D] (x1) {$x_1$};
        \node[above right=of x1] (y11) {$y_1^1$};
        \node[right=of x1] (y12) {$y_1^2$};
        \node[below right=of x1] (y13) {$y_1^3$};
        
        \node[right=of y11] (dots1) {$\cdots$};
        \node[right=of y12] (x22) {$x_2^2$};
        \node[right=of y13] (dots2) {$\cdots$};
        
        \node[above right=of x22] (y21) {$y_2^{2,1}$};
        \node[right=of x22] (y22) {$y_2^{2,2}$};
        \node[below right=of x22] (y23) {$y_2^{2,3}$};
        
        \node[right=of y21] (x31) {$\cdots$};
        \node[right=of y22] (x32) {$x_3^{2,2}$};
        \node[right=of y23] (x33) {$\cdots$};
        
        \node[above right=of x32] (y31) {$\cdots$};
        \node[right=of x32] (y32) {$\cdots$};
        \node[below right=of x32] (y33) {$\cdots$};
        
        \draw (D) -- (x1);
        \draw (x1) -- (y11);
        \draw (x1) -- (y12);
        \draw (x1) -- (y13);
        \draw (y11) -- (dots1);
        \draw (y13) -- (dots2);
        
        \draw (y12) -- (x22);
        \draw (x22) -- (y21);
        \draw (x22) -- (y22);
        \draw (x22) -- (y23);
        
        \draw (y21) -- (x31);
        \draw (y22) -- (x32);
        \draw (y23) -- (x33);
        
        \draw (x32) -- (y31);
        \draw (x32) -- (y32);
        \draw (x32) -- (y33);
    \end{tikzpicture}
    \begin{tikzpicture}[node distance=0.5cm]
        \node[draw, rectangle] (D) {$D_0$};
        \node[above right=of D] (f1) {$\hat{f}^1$};
        \node[right=of D] (f2) {$\hat{f}^2$};
        \node[below right=of D] (f3) {$\hat{f}^3$};
        
        \node[right=of f1] (x11) {$x_1$};
        \node[right=of f2] (x12) {$x_1$};
        \node[right=of f3] (x13) {$x_1$};
        
        \node[right=of x11] (y11) {$y_1^1$};
        \node[right=of x12] (y12) {$y_1^2$};
        \node[right=of x13] (y13) {$y_1^3$};
        
        \node[right=of y11] (x21) {$\cdots$};
        \node[right=of y12] (x22) {$\cdots$};
        \node[right=of y13] (x23) {$\cdots$};
        
        \draw (D) -- (f1);
        \draw (D) -- (f2);
        \draw (D) -- (f3);
        \draw (f1) -- (x11);
        \draw (f2) -- (x12);
        \draw (f3) -- (x13);
        \draw (x11) -- (y11);
        \draw (x12) -- (y12);
        \draw (x13) -- (y13);
        \draw (y11) -- (x21);
        \draw (y12) -- (x22);
        \draw (y13) -- (x23);
    \end{tikzpicture}
}
\caption{Posterior predictive sampling (left) and Pathwise sampling (right)}
\label{fig:mc_tree}
\end{figure}

When the surrogate model is a Gaussian Process (GP), the Monte Carlo method is employed to evaluate the posterior predictive distribution. In prior works, this is done via iterative sampling of the following factorized distribution: $p(y_{1:T} | x_{1:T}, D_0) = \prod_{t=1}^T p(y_t | x_t, x_{<t}, y_{<t}, D_0)$. The posterior predictive distribution at the $t$-th step, denoted as $p(y_t|x_t, x_{<t}, y_{<t}, D_0)$, can be approximated by generating $k$ samples of $y_t$ from the GP model. In general, the value of $k$ varies depending on the specific problem. At iteration $t$, suppose that we always sample $k$ samples from the posterior predictive distribution. The number of $y_t$ is $k^t$. This number quickly explodes exponentially with the length of the lookahead horizon (Figure \ref{fig:mc_tree}). The GP posterior predictive sampling process involves computing the square root of the covariance matrix, which is typically done via Cholesky decomposition. The complexity of this process is proved as $\mathcal{O}(n^3)$ for exact GP or $\mathcal{O}(m^3)$ for approximate GP where $n$ is the total number of samples in the training dataset and $m < n$ is the number of inducing samples~\cite{quinonero2005,Wilson2020}. This evidence shows that the complexity for sampling posterior predictive distribution at step $t$-th is at least $\Oc(k^t m^3)$. One variant of this procedure that can reduce the complexity is limiting the number of sampling samples for posterior predictive approximation at further lookahead steps. For instance, at each step $t > 1$, we can set $k_{t>1} = \max (k_1/2^t, 1)$, where $k_1$ is the predefined number of samples at the first lookahead step. In these cases, we can observe that $\exists \tau: \forall t > \tau, \prod_{t=1}^T k_t = K$, where $K$ is a constant. Subsequently, the complexity at step $t$-th can be reduced to $\Oc(\prod_{t=1}^T k_t m^3) = \Oc(Km^3) = \Oc(m^3)$.

To mitigate the high complexity of above sampling process, we employ the following factorization: $p(y_{1:T} | x_{1:T}, D_0) = \int p(y_{1:T} | x_{1:T}, f) p(f|D_0) \diff f = \int p(f|D_0) \prod_{t=1}^T p(y_t | x_t, f) \diff f$. The function $f$ is drawn from the prior distribution and path-wise updated via Matheron's rule. For the $h$ path, consisting of $T$ steps each, the sampling can be done with complexity $\Oc(h \times T)$. We can approximate the integral arbitrarily well with a higher $h$. The gain comes from the fact that we do not need to iteratively compute $K^{-1}_{m,m}$ as in fantasization. If we did, the complexity, with the same number of samples, would be $\Oc(h \times (T-1)^3)$. This can be done in linear complexity with respect to the number of samples. The complexity of sampling a posterior $\hat{f}$ from $p(f|D_0)$ can be considered as $\Oc(C)$, where $C$ is a constant because the number of samples in $D_0$ is unchanged. Then, computing $y_t$ for approximate posterior predictive $p(y_t|x_t,\hat{f})$ can be done by $y_t = \hat{f}(x_t)$, which has complexity of $\Oc(1)$. Using the same technique as limiting the number of sampling samples, the complexity of approximating the posterior predictive at any lookahead step is $\Oc(K)$. Thus, the total complexity at each step $t$-th is $\Oc(C + K)$. Figure \ref{fig:mc_tree} (right) visualizes the concept of this method.

\section{Details of Baselines}
\label{sec:detail_baseline}
\begin{itemize}[leftmargin=8mm]
    \item Simple Regret (SR)~\cite{Zhao2023Revisiting} measures the regret or loss in performance between the updated model and the model that would have resulted if the optimal sample had been selected for annotation during the active learning process instead.  
    \item Expected Improvement (EI)~\cite{Mockus1989BayesianAT} is used to evaluate the usefulness of candidate samples by estimating the expected gain in the performance of a model.
    \item Probability of Improvement (PI)~\cite{1964_Kushner} calculates the probability of a candidate sample improving the performance of a model compared to the current best sample.
    \item Upper Confidence Bound (UCB)~\cite{2010_Srinivas} balances exploration and exploitation by selecting candidate samples with high uncertainty and high potential for improvement based on the upper confidence bound of their predicted performance.
    \item Knowledge Gradient (KG)~\cite{2009_Frazier} quantifies the expected improvement in the objective function value resulting from evaluating a specific point. It considers the uncertainty of the model predictions and the potential benefit of obtaining additional information about the objective function.
    \item Gittins and StableGittins~\cite{Qian2024Cost} treat the BO problem as a Pandora’s Box problem and derive a Bayesian-optimal acquisition function by reinterpreting the Gittins index as a decision rule for selecting which point to evaluate next, given its cost and uncertainty.
    \item Multistep Tree (MSL)~\cite{oneshottree}, which can look up to four steps ahead, is constrained by computational costs. We reimplement this acquisition function using Pathwise sampling, enabling a lookahead horizon of up to 20 steps.
\end{itemize}

All myopic and nonmyopic baselines are modified to include the cost function as our proposed optimization objective. Specifically, their optimization objective can be formulated as 
\[
x_{t+1} = \arg \inf_{x \in \mathcal{X}} \left( -\operatorname{Acqf}(f,x) + \lambda c(x_{1:t}, x)\right).
\]

\section{Ablation Studies on Synthetic Functions}
\label{sec:ablation_synthetic}

\subsection{Varying the Observation Noise Level}
\begin{figure}[!hbt]
    \centering
    \includegraphics[width=\textwidth]{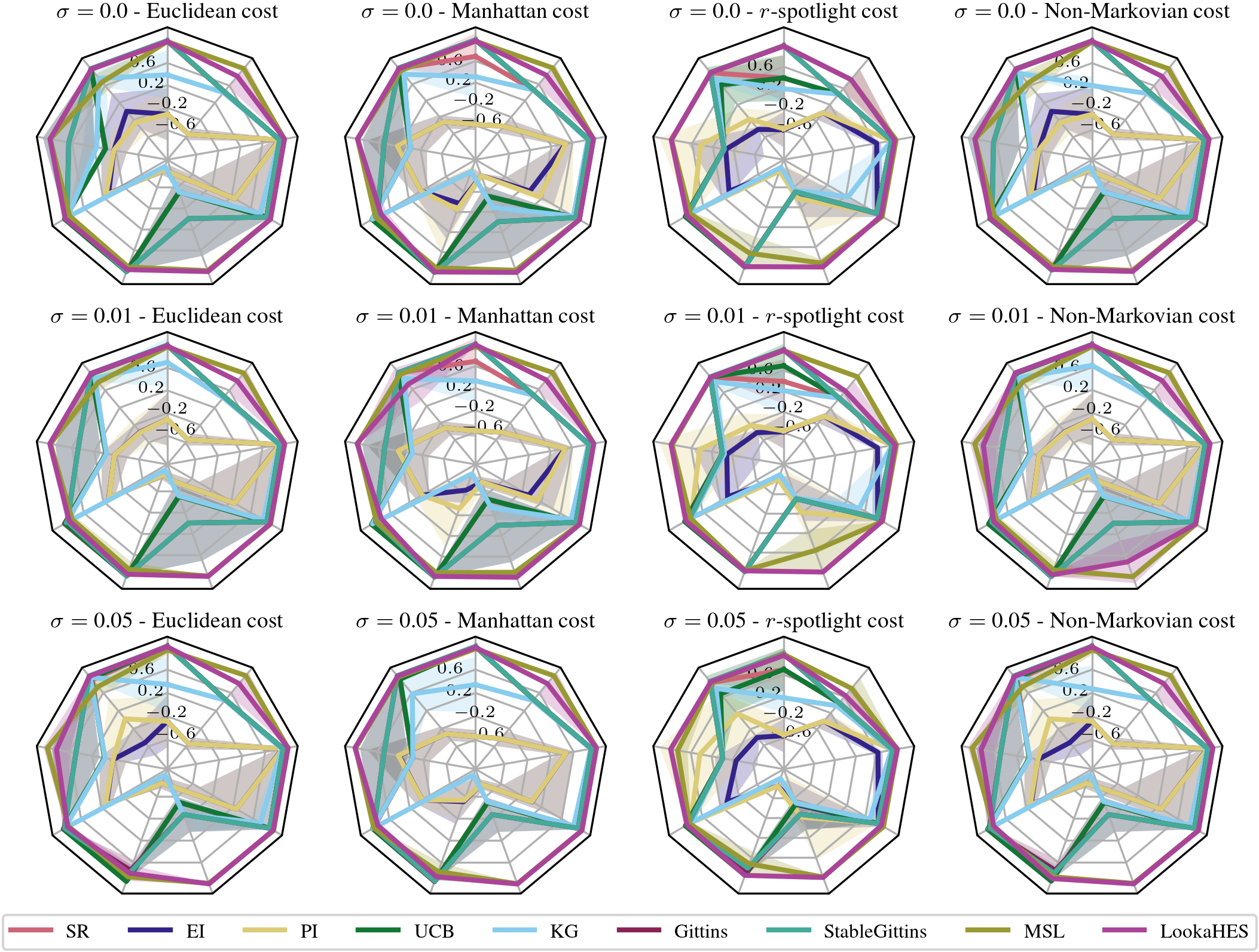}
    \caption{Final observed value. Starting from noon, counter-clockwise: Ackley, Ackley4D, Alpine, Cosine8, Hartmann, HolderTable, Levy, StyblinskiTang, SynGP. We observe that LookaHES consistently achieves the global optimum across various types of cost structures and noise levels.}
    \label{fig:final-value}
\end{figure}

To answer the \textbf{RQ3} in terms of the impact of aleatoric noise on BO methods, we performed an ablation study by varying the observation noise levels at 0\%, 1\%, and 5\%. A comprehensive comparison of LookaHES against baseline approaches was conducted across nine synthetic functions, incorporating all cost structures and noise levels. As shown in Figure~\ref{fig:final-value}, LookaHES consistently achieves superior performance across all cost structures and noise settings.

\subsection{Varying the Number of Initial Samples}
\begin{figure}[!hbt]
    \centering
    \includegraphics[width=\textwidth]{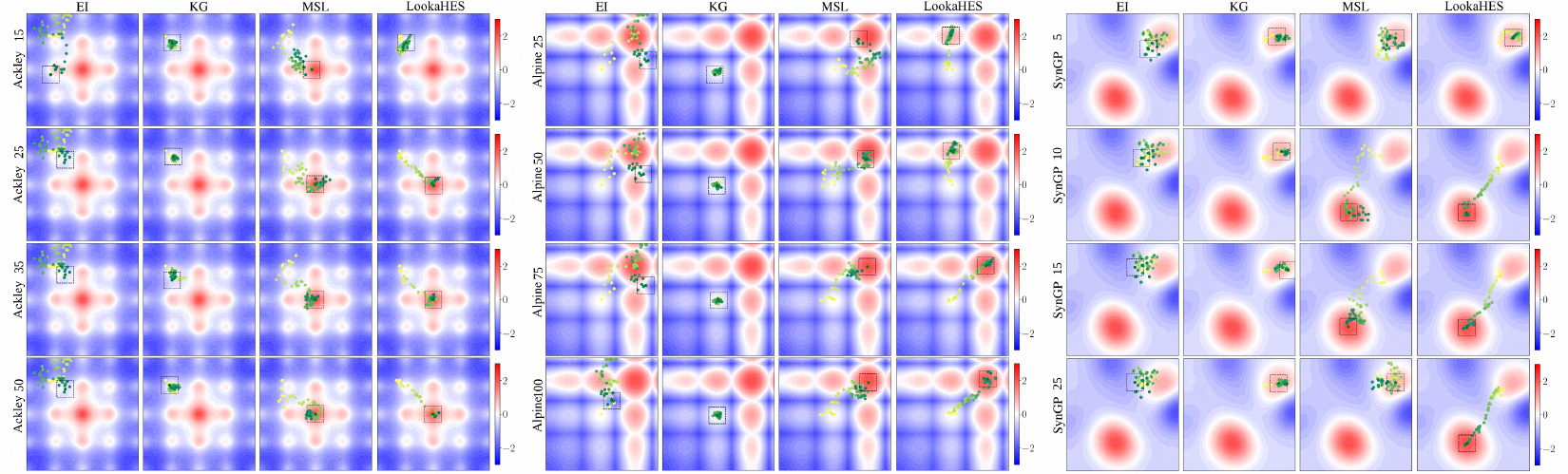}
    \caption{Comparison of performance between LookaHES and baselines with different numbers of initial samples. The yellow points indicate the starting positions, while the green points represent the final actions. From top to bottom, the Ackley function is evaluated with 15, 25, 35, and 50 initial samples; the Alpine function with 25, 50, 75, and 100 initial samples; and the SynGP function with 5, 10, 15, and 25 initial samples. With a small number of initial samples, all methods tend to fail to find the global optimum due to poor surrogate models.}
    \label{fig:abl_init_samples}
\end{figure}

In many scenarios, limited data availability at the start of an optimization process leads to a poorly constructed surrogate model, resulting in high epistemic uncertainty. Understanding the behavior of BO methods under such conditions is crucial for planning appropriate actions. Specifically, we investigated how varying the number of initial samples affects the optimization process to answer the \textbf{RQ3}. The analysis was conducted across three environments: Ackley, Alpine, and SynGP, with evaluations performed at three different levels of initial samples (Figure~\ref{fig:abl_init_samples}). According to Table~\ref{tab:abl_init_samples}, our results indicate that with fewer initial points, the GP surrogate model struggles to accurately approximate the ground-truth function, thereby increasing the likelihood of suboptimal outcomes across both myopic and nonmyopic methods.

\begin{figure}
    \centering
    \includegraphics[width=0.7\linewidth]{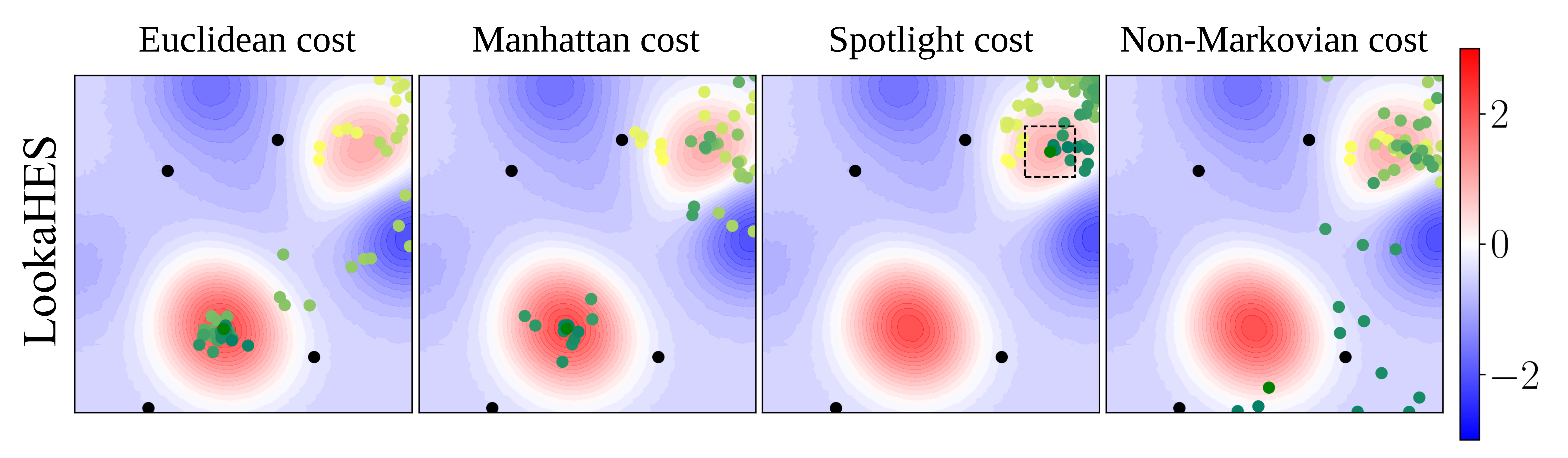}
    \caption{SynGP environment with 5 initial points and our diversity-enhanced BO method. The black points are data. The yellow points indicate the starting positions, while the green points represent the final actions.}
    \label{fig:improved_initial_samples}
\end{figure}

To address this issue, we enhance the diversity of the generated outputs and introduce a warm-up phase for the amortized network parameters during each BO iteration. Specifically, we set the $\kappa$ concentration hyperparameter of the von Mises–Fisher distribution (distribution of noises added to output) to 0, which promotes a more diverse range of outputs. For the warm-up phase, the amortized network is initialized with randomly generated data points, preventing the generated outputs from being confined to a local region. This approach ensures broader coverage of the receptive field, facilitating better exploration. The refined results are presented in Figure~\ref{fig:improved_initial_samples}, demonstrating the effectiveness of this strategy within the SynGP environment.

\begin{table*}[ht]
\centering
\caption{Comparison of performance between LookaHES and baselines with different numbers of initial samples.}
\label{tab:abl_init_samples}
\begin{tabular}{lccccc}
\toprule
\textbf{Function} & \textbf{Init Points} & \textbf{EI} & \textbf{KG} & \textbf{MSL} & \textbf{LookaHES} \\
\midrule

\multirow{4}{*}{Ackley}
& 15 & $0.276 \pm 0.490$ & $0.042 \pm 0.051$ & $0.533 \pm 0.283$ & $\mathbf{0.678 \pm 0.402}$ \\
& 25 & $-0.545 \pm 0.308$ & $0.329 \pm 0.465$ & $\mathbf{0.986 \pm 0.017}$ & $0.965 \pm 0.021$ \\
& 35 & $-0.540 \pm 0.302$ & $0.328 \pm 0.468$ & $0.943 \pm 0.027$ & $\mathbf{0.991 \pm 0.014}$ \\
& 50 & $-0.553 \pm 0.317$ & $0.164 \pm 0.289$ & $0.963 \pm 0.034$ & $\mathbf{0.993 \pm 0.013}$ \\
\midrule

\multirow{4}{*}{Alpine}
& 25 & $-0.174 \pm 0.277$ & $-0.016 \pm 0.008$ & $0.455 \pm 0.355$ & $\mathbf{0.654 \pm 0.237}$ \\
& 50 & $-0.008 \pm 0.493$ & $-0.047 \pm 0.051$ & $\mathbf{0.915 \pm 0.048}$ & $0.571 \pm 0.320$ \\
& 75 & $-0.113 \pm 0.400$ & $-0.008 \pm 0.006$ & $\mathbf{0.901 \pm 0.120}$ & $0.823 \pm 0.235$ \\
& 100 & $-0.271 \pm 0.411$ & $-0.017 \pm 0.012$ & $0.973 \pm 0.014$ & $\mathbf{0.990 \pm 0.005}$ \\
\midrule

\multirow{4}{*}{SynGP}
& 5 & $0.173 \pm 0.110$ & $0.418 \pm 0.024$ & $0.452 \pm 0.003$ & $\mathbf{0.449 \pm 0.001}$ \\
& 10 & $0.074 \pm 0.065$ & $0.416 \pm 0.050$ & $0.631 \pm 0.256$ & $\mathbf{0.810 \pm 0.254}$ \\
& 15 & $-0.001 \pm 0.172$ & $0.401 \pm 0.036$ & $0.804 \pm 0.251$ & $\mathbf{0.808 \pm 0.249}$ \\
& 25 & $-0.001 \pm 0.172$ & $0.379 \pm 0.054$ & $\mathbf{0.920 \pm 0.100}$ & $0.810 \pm 0.255$ \\
\bottomrule
\end{tabular}
\end{table*}

\subsection{Varying the Choice of Gaussian Process Kernel}
\begin{figure}[!hbt]
    \centering
    \includegraphics[width=0.48\textwidth]{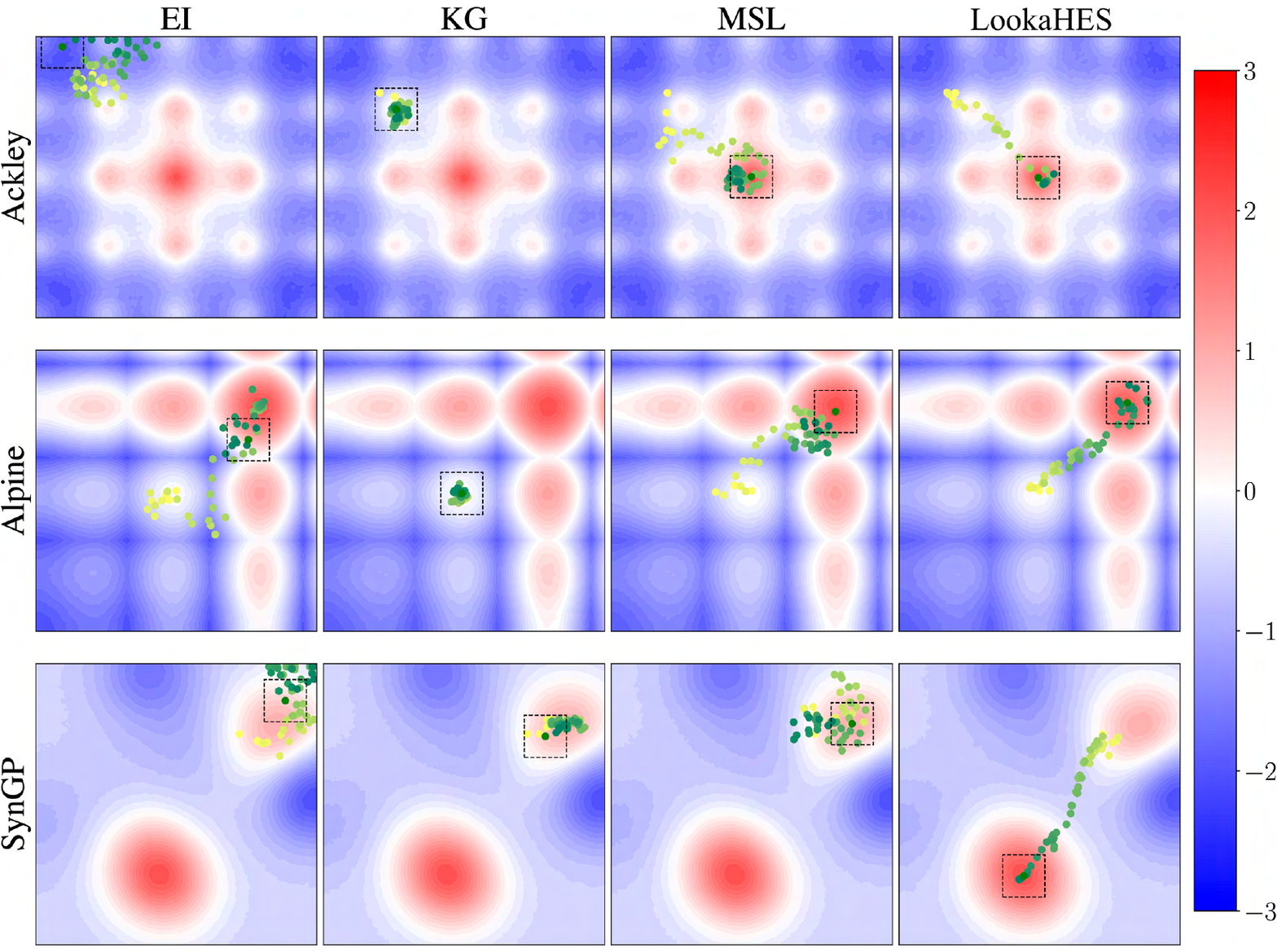}
    \includegraphics[width=0.48\textwidth]{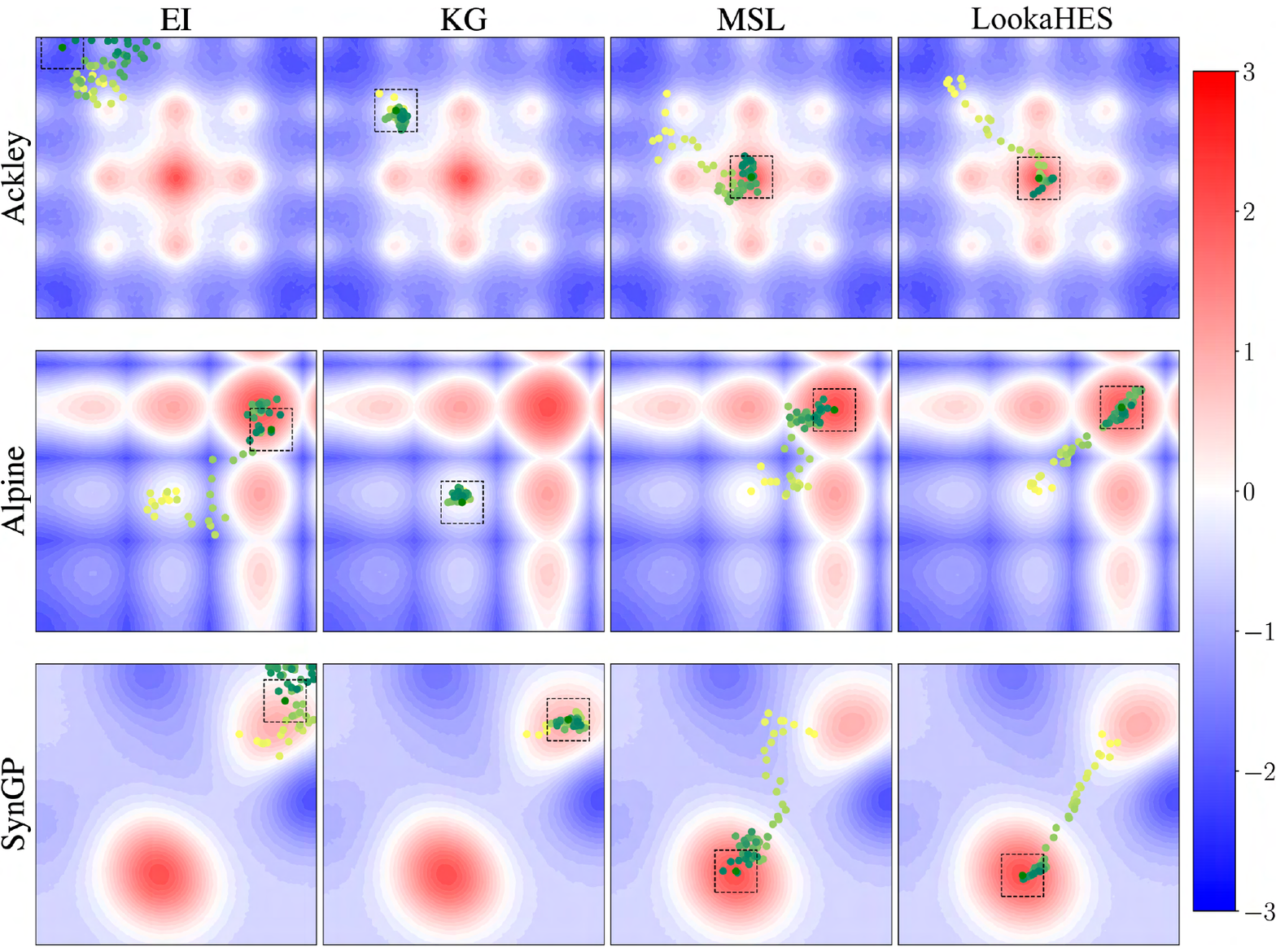}
    \caption{Comparison of performance between LookaHES and baselines with different kernels for the surrogate model (RBF on the left, and Matern on the right). The yellow points indicate the starting positions, while the green points represent the final actions. The performance of LookaHES is not affected by the choice of kernel for the surrogate model as long as the surrogate model can approximate the target function effectively.}
    \label{fig:abl_kernel}
\end{figure}

Since our surrogate models are GPs, their quality is influenced not only by the number of data points but also by the choice of kernel. In this section, we address the \textbf{RQ3}, focusing on the impact of kernel selection on GP performance. To evaluate this, we tested different kernel functions, including the Radial Basis Function (RBF) kernel and the Matérn kernel with $\nu = 1.5$, across three functions: Ackley, Alpine, and SynGP. Figure~\ref{fig:abl_kernel} visualizes the ablation results. Numerical results are presented in Table~\ref{tab:abl_kernel}. This ablation demonstrates that with any well-fitted kernel, the nonmyopic approach can achieve the global optimum.

\begin{figure}[!hbt]
    \centering
    \includegraphics[width=0.8\textwidth]{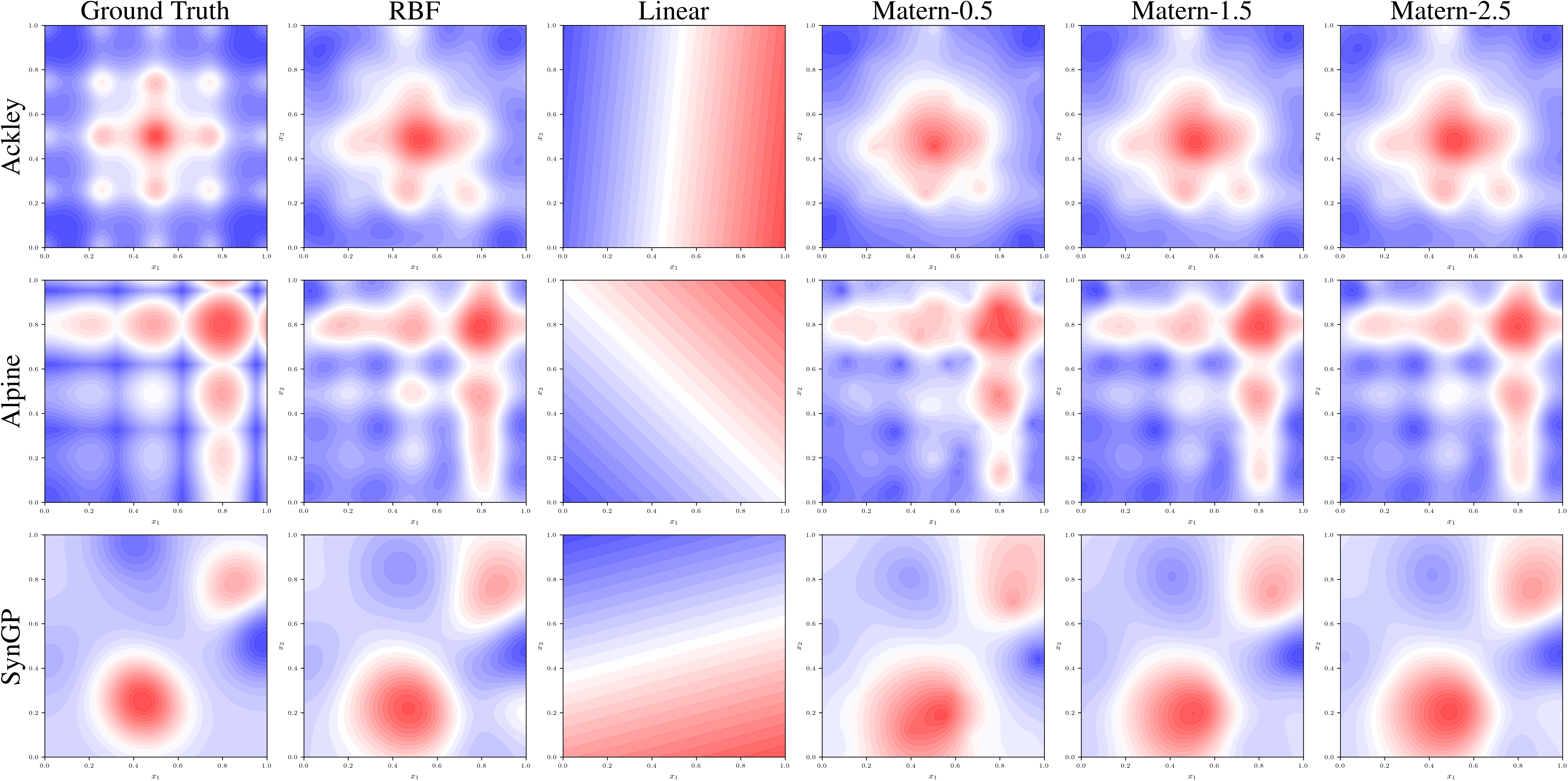}
    \caption{Comparison of posterior surfaces of different kernels on Ackley, Alpine, and SynGP function. Using Bayesian linear regression (the third column) resulted in a wrong approximation of the ground truth functions.}
    \label{fig:abl_kernel_surface}
\end{figure}

In our synthetic experiments, we do not include an ablation study on the Bayesian linear regression model, as it is unsuitable for accurately approximating the non-linear target functions. To demonstrate this limitation, we compared the posterior surface generated by Bayesian linear regression with those of other kernel-based methods, as shown in Figure~\ref{fig:abl_kernel_surface}. These results confirmed its inadequacy, leading us to exclude it from our ablation study.

\begin{table}[!ht]
\centering
\caption{Comparison of performance between LookaHES and baselines with different kernels for the surrogate model.}
\label{tab:abl_kernel}
\begin{tabular}{llcccc}
\toprule
\textbf{Kernel} & \textbf{Function} & \textbf{EI} & \textbf{KG} & \textbf{MSL} & \textbf{LookaHES} \\
\midrule

\multirow{3}{*}{RBF}
& Ackley & -0.553 $\pm$ 0.317 & 0.164 $\pm$ 0.289 & 0.963 $\pm$ 0.034 & 0.993 $\pm$ 0.013 \\
& Alpine & -0.271 $\pm$ 0.411 & -0.017 $\pm$ 0.012 & 0.973 $\pm$ 0.014 & 0.990 $\pm$ 0.005 \\
& SynGP  & -0.001 $\pm$ 0.172 & 0.379 $\pm$ 0.054 & 0.920 $\pm$ 0.100 & 0.810 $\pm$ 0.255 \\
\midrule

\multirow{3}{*}{Matern 0.5}
& Ackley & -0.553 $\pm$ 0.317 & 0.215 $\pm$ 0.544 & 0.881 $\pm$ 0.102 & -0.101 $\pm$ 0.461 \\
& Alpine & -0.059 $\pm$ 0.705 & -0.076 $\pm$ 0.047 & 0.976 $\pm$ 0.009 & 0.317 $\pm$ 0.474 \\
& SynGP  & -0.001 $\pm$ 0.172 & 0.398 $\pm$ 0.018 & 0.974 $\pm$ 0.003 & -0.039 $\pm$ 0.349 \\
\midrule

\multirow{3}{*}{Matern 1.5}
& Ackley & -0.553 $\pm$ 0.317 & 0.067 $\pm$ 0.049 & 0.969 $\pm$ 0.007 & 0.769 $\pm$ 0.289 \\
& Alpine & -0.188 $\pm$ 0.525 & -0.060 $\pm$ 0.019 & 0.988 $\pm$ 0.005 & 0.823 $\pm$ 0.236 \\
& SynGP  & -0.001 $\pm$ 0.172 & 0.400 $\pm$ 0.044 & 0.989 $\pm$ 0.003 & 0.630 $\pm$ 0.255 \\
\midrule

\multirow{3}{*}{Matern 2.5}
& Ackley & -0.553 $\pm$ 0.317 & 0.067 $\pm$ 0.049 & 0.969 $\pm$ 0.007 & 0.769 $\pm$ 0.289 \\
& Alpine & -0.188 $\pm$ 0.525 & -0.060 $\pm$ 0.019 & 0.988 $\pm$ 0.005 & 0.823 $\pm$ 0.236 \\
& SynGP  & -0.001 $\pm$ 0.172 & 0.400 $\pm$ 0.044 & 0.989 $\pm$ 0.003 & 0.630 $\pm$ 0.255 \\
\bottomrule
\end{tabular}
\end{table}

\subsection{Varying the Length of the Lookahead Horizon}
\begin{figure}[!htb]
    \centering
    \includegraphics[width=\textwidth]{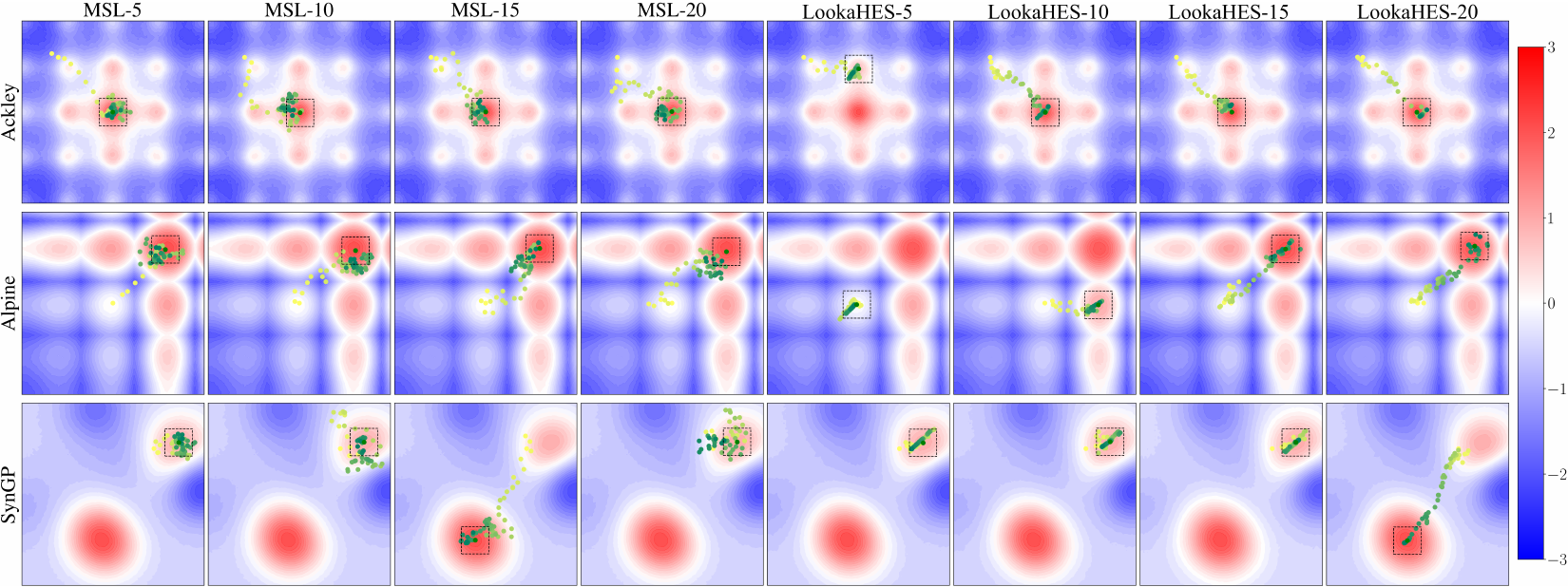}
    \caption{Comparison of LookaHES and nonmyopic baseline at 5, 10, 15, and 20 lookahead steps. The yellow points indicate the starting positions, while the green points represent the final actions. With fewer lookahead steps, nonmyopic methods tend to fail to find the global optimum, demonstrating the benefit of having a longer lookahead horizon.}
    \label{fig:abl_lookahead}
\end{figure}

To answer \textbf{RQ3} on the benefit of a large lookahead horizon, we included experimental results on the ablation of the number of lookahead steps in Figure~\ref{fig:abl_lookahead} and Table~\ref{tab:abl_lookahead}. These results illustrate the relationship between the number of lookahead steps and the robustness of the optimization, providing insights into how the performance of our approach varies with different horizon lengths. Specifically, with a smaller lookahead horizon, the probability of being trapped by local optima increases, leading to suboptimal optimization in all nonmyopic methods.

The experiments conducted across various scenarios highlight the robustness and effectiveness of our proposed method in handling different challenges in Bayesian optimization. We observed that varying observation noise levels (0\%, 1\%, and 5\%) had minimal impact on the performance of LookaHES compared to baseline approaches, with LookaHES consistently achieving the global optimum across all cost structures and noise settings. Additionally, we found that limited initial samples led to suboptimal performance due to high epistemic uncertainty, but introducing a diversity-enhancing strategy and a warm-up phase improved the results. Kernel selection also played a crucial role, and LookaHES demonstrated strong performance regardless of the kernel choice as long as the surrogate model could approximate the target function effectively. Finally, our results show that incorporating a larger lookahead horizon significantly improves the optimization process, reducing the likelihood of being trapped in local optima.

\begin{table}[!htb]
\centering
\caption{Comparison of LookaHES and nonmyopic baseline at 5, 10, 15, and 20 lookahead steps.}
\label{tab:abl_lookahead}
\begin{tabular}{llcccc}
\toprule
\textbf{Function} & \textbf{Method} & \textbf{5} & \textbf{10} & \textbf{15} & \textbf{20} \\
\midrule

\multirow{2}{*}{Ackley}
& MSL      & 0.992 $\pm$ 0.013 & 0.986 $\pm$ 0.012 & 0.983 $\pm$ 0.015 & 0.963 $\pm$ 0.034 \\
& LookaHES & 0.979 $\pm$ 0.018 & 0.991 $\pm$ 0.013 & 0.995 $\pm$ 0.009 & 0.993 $\pm$ 0.013 \\
\midrule

\multirow{2}{*}{Alpine}
& MSL      & 0.983 $\pm$ 0.008 & 0.987 $\pm$ 0.008 & 0.947 $\pm$ 0.038 & 0.973 $\pm$ 0.014 \\
& LookaHES & 0.990 $\pm$ 0.006 & 0.989 $\pm$ 0.005 & 0.990 $\pm$ 0.006 & 0.990 $\pm$ 0.005 \\
\midrule

\multirow{2}{*}{SynGP}
& MSL      & 0.452 $\pm$ 0.003 & 0.630 $\pm$ 0.253 & 0.988 $\pm$ 0.004 & 0.920 $\pm$ 0.100 \\
& LookaHES & 0.452 $\pm$ 0.003 & 0.452 $\pm$ 0.003 & 0.452 $\pm$ 0.003 & 0.810 $\pm$ 0.255 \\
\bottomrule
\end{tabular}
\end{table}

\begin{mybox}
\textbf{\textit{Summary:}} 
Our proposed method demonstrated robustness to aleatoric noise, maintaining strong performance even with a 5\% noise level. High epistemic uncertainty from limited initial samples hindered performance, but strategies like diversity enhancement and warm-up phases mitigated this issue. Additionally, effective surrogate model design and a larger lookahead horizon were crucial, enhancing optimization by avoiding local optima and improving convergence.
\end{mybox}

\subsection{Varying the Acquisition Function Hyperparameters}
\begin{sidewaysfigure}[!hbpt]
    \centering
    \includegraphics[width=\textwidth]{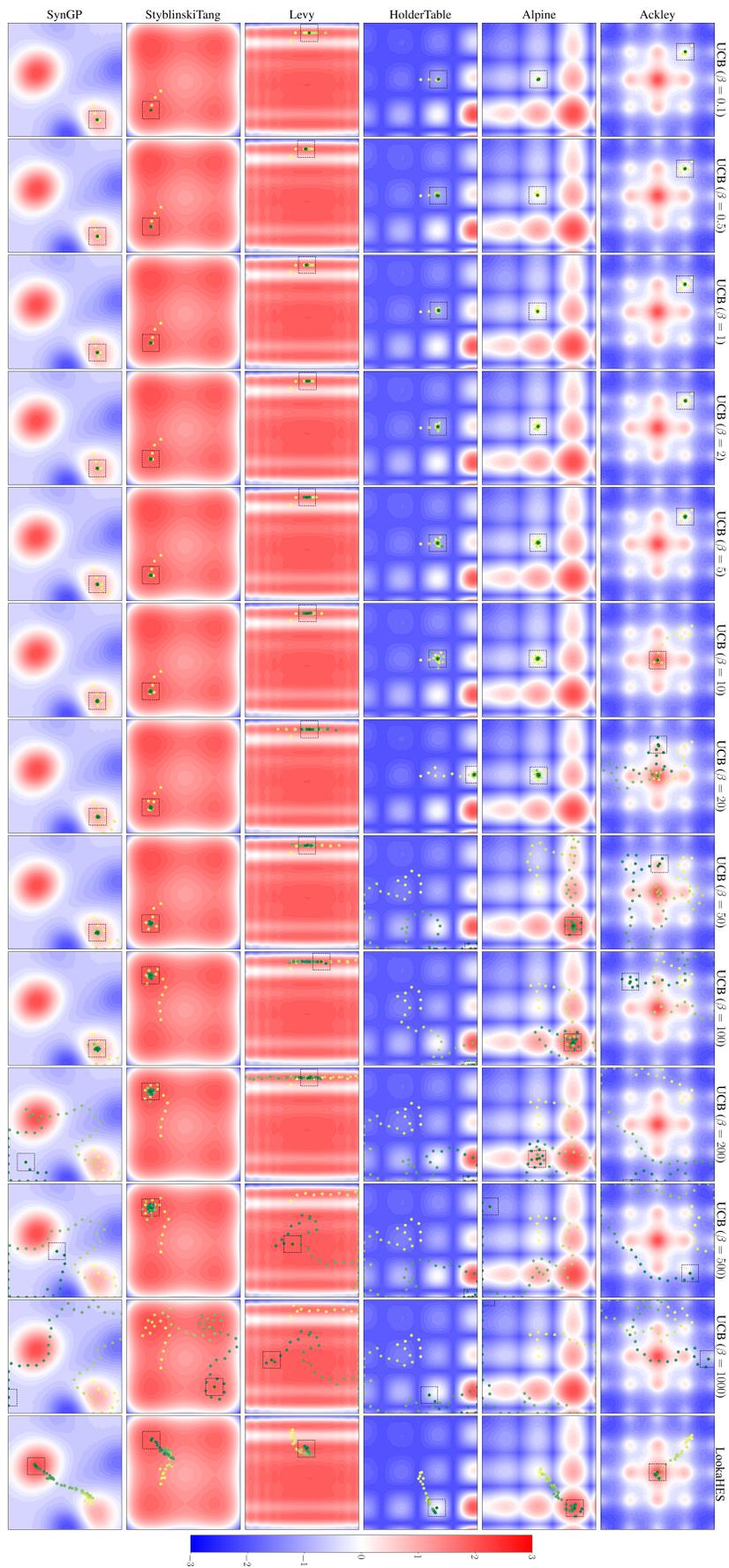}
    \caption{Visualization of queries across BO iterations with the setting of $\sigma = 0.0$ and r-spotlight cost. The yellow points indicate the starting positions, while the green points represent the final actions. With an appropriate $\beta$, the UCB can achieve the global optimum as ours.}
    \label{fig:viz_ucb}
\end{sidewaysfigure}

Myopic acquisition functions, such as UCB, rely on ``optimism" during optimization. This means they prioritize exploration with the expectation that querying enough points may eventually uncover the optimum. In this section, we investigate whether this ``optimism" can outperform lookahead methods, addressing \textbf{RQ4}. In the case of the UCB acquisition function, the degree of optimism is controlled by the $\beta$ hyperparameter: smaller values of $\beta$ emphasize exploitation, while larger values encourage exploration. With sufficiently large $\beta$, the standard deviation term dominates the mean term, leading to decisions driven by the most uncertain areas. Additionally, the PI acquisition function has a $\tau$ parameter to control optimism, with smaller values encouraging exploration. To further illustrate the impact of $\beta$ and $\tau$, we conducted additional experiments with $\beta$ values ranging from $0.1$ to $1000$, $\tau$ values ranging from $0.1$ to $20$ on nine synthetic functions. Other myopic functions, including SR, EI, and KG, do not have a mechanism to handle optimism. In Figure~\ref{fig:viz_ucb}, we highlight the behavior of UCB when increasing $\beta$.

We also provide the value of the final action, normalized to range from -1 to 1, where -1 represents the worst outcome and 1 is the best in Table~\ref{tab:ucb_p1}. These empirical results further illustrate that the large $\beta$ value can encourage the decision-maker to make queries that highly prioritize exploration. As illustrated in the above figure and table, such exploration is typically myopic and unplanned, and consequently, the decision maker typically misses the global optima or overexplores the un-promising region. We also want to note that in our experiment, no single $\beta$ outperformed others in all settings: for example, $\beta = 10$ works well for Ackley, but does not work for other functions. Indeed, choosing the value of $\beta$ for UCB before running the online experiment is nontrivial in practice.

\begin{table}[!htbp]
    \centering
\caption{Comparison of the final action value of LookaHES with 20-step lookahead and UCB with various $\beta$ values}
\label{tab:ucb_p1}
\resizebox{\textwidth}{!}{
\begin{tabular}{lccccccccc}
\toprule
\textbf{Method} & \textbf{Ackley} & \textbf{Ackley4D} & \textbf{Alpine} & \textbf{Cosine8} & \textbf{Hartmann} & \textbf{HolderTable} & \textbf{Levy} & \textbf{StyblinskiTang} & \textbf{SynGP} \\
\midrule 
\text{LookaHES} & $0.99 \pm 0.01$ & $0.97 \pm 0.02$ & $\mathbf{0.99 \pm 0.01}$ & $0.93 \pm 0.01$ & $0.96 \pm 0.03$ & $0.05 \pm 0.08$ & $\mathbf{0.95 \pm 0.0}$ & $\mathbf{1.0 \pm 0.0}$ & $\mathbf{0.81 \pm 0.26}$ \\
\midrule
\text{UCB} ($\beta = 0.1$) & $0.7 \pm 0.4$ & $0.68 \pm 0.44$ & $-0.01 \pm 0.01$ & $0.96 \pm 0.01$ & $0.95 \pm 0.02$ & $-0.49 \pm 0.01$ & $0.87 \pm 0.0$ & $0.91 \pm 0.0$ & $0.45 \pm 0.01$ \\
\text{UCB} ($\beta = 0.5$) & $0.4 \pm 0.41$ & $0.68 \pm 0.44$ & $-0.01 \pm 0.01$ & $\mathbf{0.97 \pm 0.01}$ & $0.94 \pm 0.04$ & $-0.5 \pm 0.01$ & $0.87 \pm 0.0$ & $0.91 \pm 0.0$ & $0.45 \pm 0.01$ \\ 
\text{UCB} ($\beta = 1$) & $0.4 \pm 0.41$ & $0.67 \pm 0.43$ & $-0.01 \pm 0.01$ & $0.96 \pm 0.02$ & $0.95 \pm 0.03$ & $-0.49 \pm 0.01$ & $0.87 \pm 0.0$ & $0.91 \pm 0.0$ & $0.45 \pm 0.01$ \\
\text{UCB} ($\beta = 2$) & $0.4 \pm 0.41$ & $0.68 \pm 0.44$ & $-0.01 \pm 0.01$ & $0.97 \pm 0.02$ & $0.95 \pm 0.03$ & $-0.49 \pm 0.01$ & $0.87 \pm 0.0$ & $0.91 \pm 0.0$ & $0.45 \pm 0.01$ \\
\text{UCB} ($\beta = 5$) & $0.7 \pm 0.43$ & $0.98 \pm 0.01$ & $-0.01 \pm 0.01$ & $0.96 \pm 0.03$ & $\mathbf{0.97 \pm 0.02}$ & $-0.49 \pm 0.01$ & $0.87 \pm 0.0$ & $0.91 \pm 0.0$ & $0.45 \pm 0.01$ \\
\text{UCB} ($\beta = 10$) & $\mathbf{1.0 \pm 0.01}$ & $0.98 \pm 0.01$ & $-0.01 \pm 0.01$ & $0.97 \pm 0.02$ & $0.96 \pm 0.03$ & $-0.28 \pm 0.32$ & $0.87 \pm 0.0$ & $0.91 \pm 0.0$ & $0.45 \pm 0.01$ \\
\text{UCB} ($\beta = 20$) & $0.71 \pm 0.35$ & $0.87 \pm 0.17$ & $0.15 \pm 0.24$ & $0.97 \pm 0.04$ & $0.97 \pm 0.03$ & $-0.03 \pm 0.34$ & $0.87 \pm 0.0$ & $0.91 \pm 0.0$ & $0.45 \pm 0.01$ \\
\text{UCB} ($\beta = 50$) & $0.55 \pm 0.65$ & $\mathbf{0.99 \pm 0.01}$ & $0.73 \pm 0.36$ & $0.88 \pm 0.04$ & $0.94 \pm 0.03$ & $0.81 \pm 0.6$ & $0.87 \pm 0.0$ & $0.91 \pm 0.0$ & $0.45 \pm 0.01$ \\
\text{UCB} ($\beta = 100$) & $0.45 \pm 0.41$ & $0.47 \pm 0.33$ & $0.92 \pm 0.09$ & $0.71 \pm 0.06$ & $0.95 \pm 0.03$ & $0.9 \pm 0.81$ & $0.86 \pm 0.01$ & $0.94 \pm 0.04$ & $0.45 \pm 0.01$ \\
\text{UCB} ($\beta = 200$) & $-0.34 \pm 0.38$ & $-0.72 \pm 0.06$ & $0.15 \pm 0.46$ & $0.62 \pm 0.07$ & $0.81 \pm 0.12$ & $\mathbf{0.76 \pm 0.23}$ & $0.93 \pm 0.05$ & $0.93 \pm 0.1$ & $0.22 \pm 0.32$ \\
\text{UCB} ($\beta = 500$) & $-0.12 \pm 0.13$ & $-0.33 \pm 0.3$ & $-0.43 \pm 0.48$ & $0.13 \pm 0.18$ & $-0.02 \pm 0.61$ & $0.27 \pm 1.1$ & $0.82 \pm 0.12$ & $0.96 \pm 0.06$ & $0.61 \pm 0.53$ \\
\text{UCB} ($\beta = 1000$) & $0.06 \pm 0.5$ & $-0.6 \pm 0.24$ & $-0.2 \pm 0.66$ & $-0.18 \pm 0.1$ & $-0.52 \pm 0.43$ & $-0.41 \pm 0.32$ & $0.84 \pm 0.06$ & $0.92 \pm 0.07$ & $0.5 \pm 0.53$ \\ 
\midrule
\text{PI ($\tau=0.1$)} & $-0.55 \pm 0.32$ & $-0.40 \pm 0.06$ & $-0.05 \pm 0.72$ & $0.12 \pm 0.25$ & $-0.85 \pm 0.04$ & $-0.23 \pm 0.84$ & $0.45 \pm 0.77$ & $0.72 \pm 0.13$ & $-0.00 \pm 0.17$ \\
\text{PI ($\tau=0.5$)} & $-0.51 \pm 0.27$ & $-0.12 \pm 0.38$ & $0.38 \pm 0.75$ & $0.71 \pm 0.02$ & $-0.85 \pm 0.04$ & $-0.30 \pm 0.74$ & $0.84 \pm 0.02$ & $0.88 \pm 0.02$ & $-0.00 \pm 0.17$ \\
\text{PI ($\tau=1$)}   & $0.21 \pm 0.55$  & $0.48 \pm 0.05$  & $-0.13 \pm 0.02$ & $0.93 \pm 0.02$ & $0.87 \pm 0.10$  & $-0.59 \pm 0.07$ & $0.84 \pm 0.02$ & $0.91 \pm 0.00$ & $0.38 \pm 0.04$  \\
\text{PI ($\tau=2$)}   & $0.40 \pm 0.41$  & $0.64 \pm 0.42$  & $-0.10 \pm 0.05$ & $0.93 \pm 0.03$ & $0.96 \pm 0.03$  & $-0.60 \pm 0.05$ & $0.87 \pm 0.00$ & $0.91 \pm 0.00$ & $0.45 \pm 0.01$  \\
\text{PI ($\tau=5$)}   & $0.70 \pm 0.41$  & $0.66 \pm 0.43$  & $-0.01 \pm 0.01$ & $0.96 \pm 0.01$ & $0.94 \pm 0.04$  & $-0.50 \pm 0.01$ & $0.87 \pm 0.00$ & $0.91 \pm 0.00$ & $0.45 \pm 0.01$  \\
\text{PI ($\tau=10$)}  & $0.70 \pm 0.40$  & $0.97 \pm 0.01$  & $-0.01 \pm 0.01$ & $0.96 \pm 0.01$ & $0.94 \pm 0.03$  & $-0.49 \pm 0.01$ & $0.87 \pm 0.00$ & $0.91 \pm 0.00$ & $0.45 \pm 0.01$  \\
\text{PI ($\tau=20$)}  & $0.70 \pm 0.40$  & $0.95 \pm 0.05$  & $-0.01 \pm 0.01$ & $0.95 \pm 0.01$ & $0.93 \pm 0.01$  & $-0.49 \pm 0.01$ & $0.87 \pm 0.00$ & $0.91 \pm 0.00$ & $0.45 \pm 0.01$  \\
\bottomrule 
\end{tabular}
}
\end{table}

Our experiments demonstrate that while large values of the $\beta$ hyperparameter in the UCB acquisition function prioritize exploration, they often lead to myopic and unplanned exploration, causing the decision-maker to miss the global optimum or over-explore unpromising regions. The performance of UCB varies across different functions, with no single $\beta$ value outperforming others universally, highlighting the challenge of selecting an optimal $\beta$ before running the experiment. This underscores the complexity of using optimism in myopic methods compared to more structured lookahead approaches.

\begin{mybox}
\textbf{\textit{Summary:}} 
Optimism in myopic methods, such as using a large \(\beta\) in the UCB acquisition function, can lead to unplanned exploration and suboptimal performance, as it risks missing the global optimum or over-exploring unpromising areas. While this optimism may occasionally benefit specific scenarios, its lack of consistency across functions limits its broad applicability to real-world problems compared to more structured, nonmyopic approaches.
\end{mybox}

\subsection{Synthetic Functions with Discrete Input} 
In many practical scenarios, data domains exhibit discrete attributes, such as those encountered in natural language processing or chemical molecular structures. To answer the \textbf{RQ2} and demonstrate the efficiency of LookaHES within discrete spaces, we conducted an additional experiment. In this experiment, we utilized the SynGP environment. However, in this case, we discretized the domain of each dimension into \( C \) categories, such that the design variable belongs to the set \( \Cbb^d \), where \( d \) represents the number of design dimensions. The categorical variables are represented in a one-hot encoding format. At time step \( t \), the design variable \( x_t \) is a matrix of size \( C \times d \), with \( x_t = [x_t^i]_{i=1}^d \), where \( x_t^i \in \Cbb \) denotes the one-hot encoded vector for the \( i \)-th dimension.

To adapt our variational network to this discrete domain, we modify its architecture as follows. To compute the representation \( x_t' \) of the design variable for the variational network, each dimension \( x_t^i \) is passed through a linear transformation, followed by a sum-pooling operation across dimensions: $x_t' = \sum_{i=1}^d W x_t^i$.
The output of the variational network is a probability vector corresponding to a categorical distribution. To ensure gradient propagation during optimization, we sample the most likely category using the Straight-through reparameterization technique \cite{Bengio2013EstimatingOP}.

The MSL acquisition function in discrete domains does not benefit from gradient-based optimization methods to minimize the loss function. As a result, MSL is implemented using a multistep combinatorial search in this experiment. This approach significantly increases computational complexity, especially when dealing with a large number of categories. For instance, in the SynGP environment, where an input variable has 20 categories, and the MSL algorithm is configured with \( L = 4 \) lookahead steps, the number of combinations to explore is given by \( (C^d)^L = (20^2)^4 = 25.6 \times 10^9 \), leading to an exponentially large search space. Given the constraints of limited computational resources, we resort to a random search with a budget of 2000 possible design variable configurations.

\begin{figure*}[t]\centering
    \includegraphics[width=\textwidth]{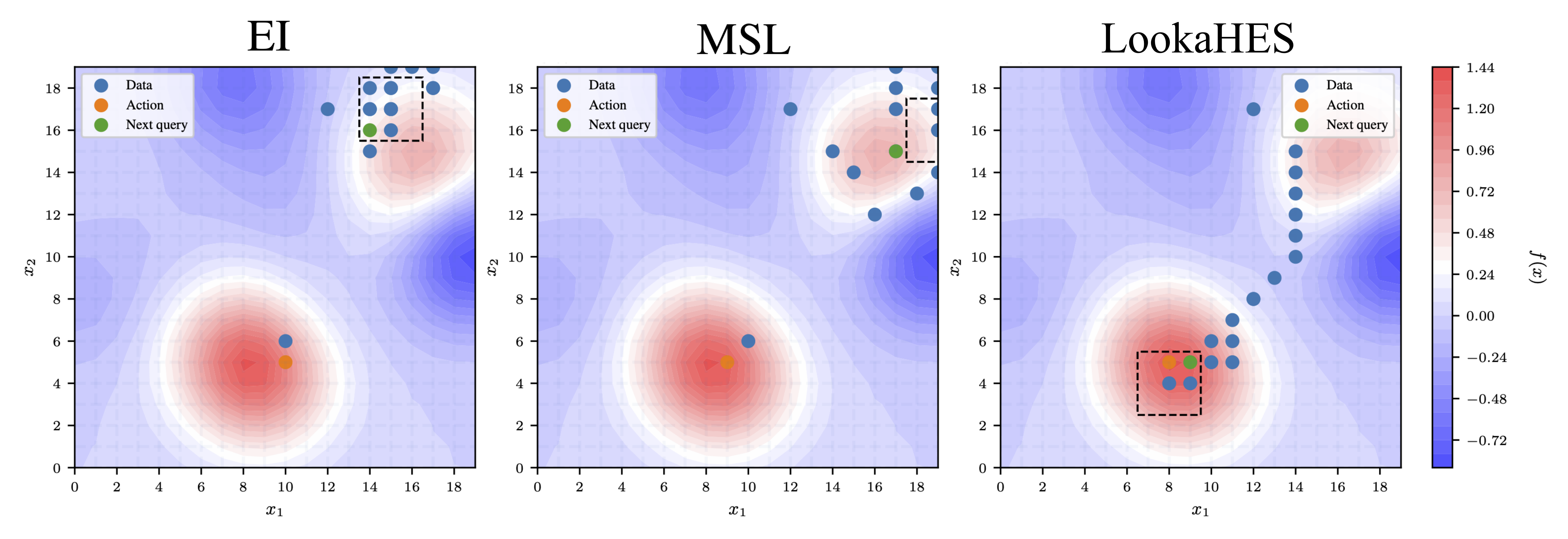}
    \caption{Results of discrete setting on SynGP. From left to right: EI, MSL, LookaHES}
    \label{fig:discrete}
\end{figure*}

Figure~\ref{fig:discrete} presents a comparative analysis of the performance of the myopic acquisition functions EI, MSL, and our proposed acquisition function within the discrete SynGP after 18 BO iterations. As shown, LookaHES outperforms the other acquisition functions. This experiment thus highlights the robustness of our approach in addressing optimization problems in discrete domains.

\subsection{Ablation of Cost Noises}
In the real world, sometimes we can not compute the cost function exactly. In other words, the cost function is only an approximation of real-world cost. We perform ablation studies on the cost functions with 5\% noise. The results are presented in Table~\ref{tab:abl_cost_noise}.

\begin{table}[ht]
\centering
\caption{Comparison of optimization methods across environment and cost function with 5\% noise.}
\label{tab:abl_cost_noise}
\begin{tabular}{llcccc}
\toprule
\textbf{Environment} & \textbf{Cost Function} & \textbf{EI} & \textbf{KG} & \textbf{MSL} & \textbf{LookaHES} \\
\midrule
\multirow{3}{*}{Ackley} & Euclidean &  $-0.031 \pm 0.441$ &   $0.980 \pm 0.003$ &  $0.966 \pm 0.012$ &  $\mathbf{0.998 \pm 0.011}$ \\
      & Manhattan &  $-0.303 \pm 0.259$ &   $0.683 \pm 0.437$ &  $0.966 \pm 0.035$ &  $\mathbf{0.993 \pm 0.010}$ \\
      & Non-Markovian &  $-0.031 \pm 0.441$ &   $0.982 \pm 0.013$ &  $0.957 \pm 0.031$ &  $\mathbf{0.995 \pm 0.011}$ \\
\cline{1-6}
\multirow{3}{*}{Alpine} & Euclidean &  $-0.352 \pm 0.257$ &   $0.325 \pm 0.476$ &  $\mathbf{0.953 \pm 0.027}$ &  $0.947 \pm 0.055$ \\
      & Manhattan &  $-0.200 \pm 0.108$ &   $0.158 \pm 0.236$ &  $0.964 \pm 0.007$ &  $\mathbf{0.990 \pm 0.006}$ \\
      & Non-Markovian &  $-0.352 \pm 0.257$ &  $-0.008 \pm 0.006$ &  $0.954 \pm 0.042$ &  $\mathbf{0.990 \pm 0.006}$ \\
\cline{1-6}
\multirow{3}{*}{SynGP} & Euclidean &  $-0.349 \pm 0.292$ &   $0.452 \pm 0.003$ &  $\mathbf{0.984 \pm 0.001}$ &  $0.810 \pm 0.256$ \\
      & Manhattan &  $-0.243 \pm 0.238$ &   $0.452 \pm 0.003$ &  $\mathbf{0.988 \pm 0.003}$ &  $0.810 \pm 0.251$ \\
      & Non-Markovian &  $-0.349 \pm 0.292$ &   $0.452 \pm 0.003$ &  $0.988 \pm 0.002$ &  $\mathbf{0.989 \pm 0.003}$ \\
\bottomrule
\end{tabular}
\end{table}

\subsection{High-Dimensional Experiments}

We further evaluate our method in high-dimensional settings using the Ackley function in 20 and 50 dimensions. For these environments, we consider three representative cost functions: \emph{Euclidean}, \emph{spotlight}, and \emph{non-Markovian}. We exclude the Manhattan cost in high-dimensional experiments, as it is rarely used in such regimes: distances under the $\ell_1$ metric tend to concentrate as dimensionality increases, rendering them less informative and often leading to degraded performance in Bayesian optimization methods. We also omit Knowledge Gradient from these experiments. In high-dimensional settings, KG becomes computationally intractable and consistently results in out-of-memory errors on an NVIDIA A100 GPU with 80\,GB of memory.

Table~\ref{tab:high_dim_results} reports the results across all high-dimensional configurations (20D and 50D). Overall, our method achieves performance that is comparable to, and in several cases matches, the strongest baseline in each setting, across all considered cost functions and dimensionalities. This demonstrates that our approach remains robust and competitive even as the dimensionality of the search space increases substantially.

\begin{table}[!ht]
\centering
\caption{Performance on high-dimensional Ackley function with 5\% observation noise under different cost functions}
\label{tab:high_dim_results}
\resizebox{\textwidth}{!}{
\begin{tabular}{lcccccc}
\toprule
\textbf{Setting (Function, Cost)} & \textbf{SR} & \textbf{EI} & \textbf{PI} & \textbf{UCB} & \textbf{MSL} & \textbf{LookaHES} \\
\midrule
Ackley20D, Euclidean 
& $0.83 \pm 0.04$ 
& $-0.69 \pm 0.00$ 
& $-0.19 \pm 0.40$ 
& $0.83 \pm 0.04$ 
& $0.64 \pm 0.06$ 
& $0.80 \pm 0.05$ \\
Ackley20D, Spotlight 
& $0.76 \pm 0.02$ 
& $-0.70 \pm 0.09$ 
& $0.01 \pm 0.44$ 
& $0.77 \pm 0.05$ 
& $0.80 \pm 0.01$ 
& $0.89 \pm 0.08$ \\
Ackley20D, Non-Markovian 
& $0.86 \pm 0.01$ 
& $-0.40 \pm 0.21$ 
& $-0.19 \pm 0.40$ 
& $0.83 \pm 0.04$ 
& $0.64 \pm 0.06$ 
& $0.85 \pm 0.08$ \\
Ackley50D, Euclidean 
& $0.84 \pm 0.02$ 
& $-0.56 \pm 0.12$ 
& $-0.56 \pm 0.00$ 
& $0.83 \pm 0.03$ 
& $0.43 \pm 0.08$ 
& $0.71 \pm 0.04$ \\
Ackley50D, Spotlight 
& $-0.27 \pm 0.21$ 
& $-0.72 \pm 0.06$ 
& $-0.79 \pm 0.06$ 
& $-0.11 \pm 0.02$ 
& $0.81 \pm 0.01$ 
& $0.74 \pm 0.03$ \\ 
Ackley50D, Non-Markovian
& $0.83 \pm 0.03$ 
& $-0.57 \pm 0.00$ 
& $-0.61 \pm 0.07$ 
& $0.83 \pm 0.03$ 
& $0.43 \pm 0.08$ 
& $0.57 \pm 0.16$ \\
\bottomrule
\end{tabular}
}
\end{table}

\section{Details of Night Light Experiments}
\label{sec:nightlight}
\begin{figure}[!htb]
    \centering
    \includegraphics[width=\linewidth]{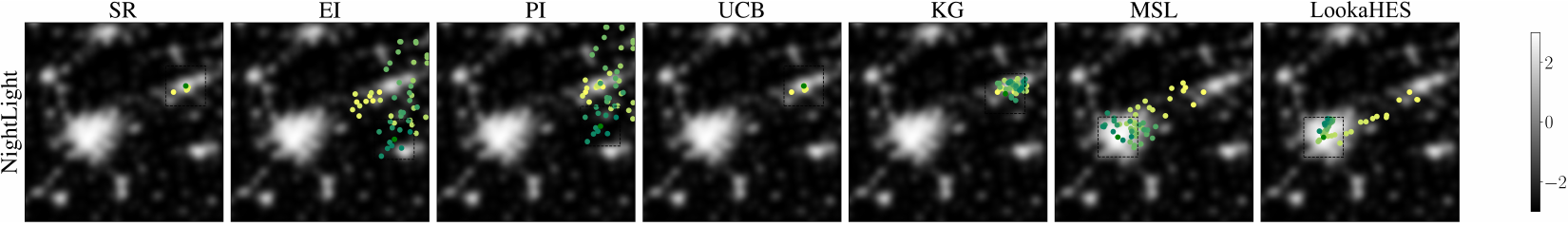}
    \includegraphics[width=\linewidth]{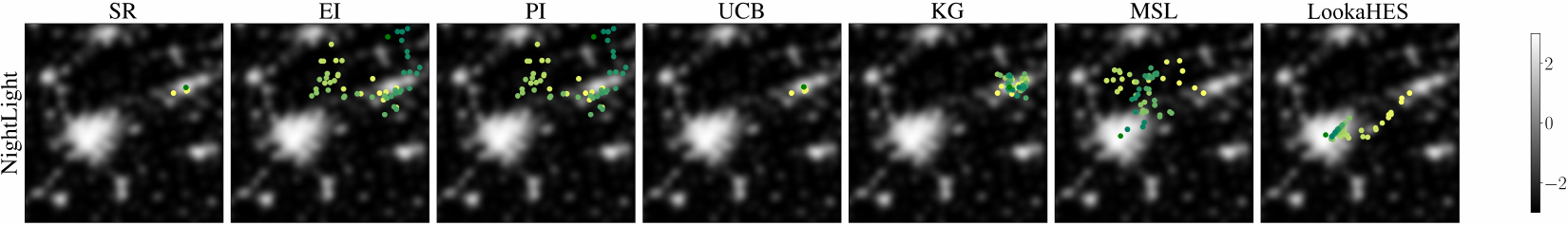}
    \caption{Visualization of different methods on NASA night light images in the case of spotlight cost (top row) and Euclidean cost (bottom row)}
    \label{fig:viz_nightlight}
\end{figure}

To demonstrate the applicability of LookaHES in real-world continuous environments, we conducted experiments on human travel optimization within a 2D continuous domain. This experiment addresses \textbf{RQ1} in the context of continuous domains. Specifically, we used a 2016 grayscale image of night lights in Georgia and South Carolina, sourced from NASA's Earth Observatory, with a resolution of $1000 \times 1000$ pixels. The data is online accessible at \url{https://earthobservatory.nasa.gov/features/NightLights}. To facilitate the optimization of the GP surrogate model and avoid numerical issues due to image noise, we applied a stack blur with a radius of 50 to the image. The pixel values, ranging from 0 to 255, were normalized to a range of $-3$ to $3$. The image width and height were normalized to a range of $0$ to $1$. We apply LookaHES and baselines with spotlight cost ($r=0.1$) and Euclidean cost. Figures~\ref{fig:viz_nightlight} show the results of LookaHES and baselines on the spotlight and Euclidean cost, respectively. In this environment, nonmyopic methods demonstrated their advantage in lookahead capability. Notably, LookaHES showed its effectiveness in directly reaching the global optimum, rather than querying around sub-optimal locations before approaching the global optimum as MSL. Table~\ref{tab:nightlight} presents the numerical results of this experiment with three different seeds.

\begin{table}[htbp]
    \centering
    \caption{Values of Final Action in The NightLight Environment. All Values Are Scaled To $(-1, 1)$.}
    \label{tab:nightlight}
    \begin{tabular}{c|ccccccc}
    \toprule
        \textbf{Cost} & \textbf{SR} & \textbf{EI} & \textbf{PI} & \textbf{UCB} & \textbf{KG} & \textbf{MSL} & \textbf{LookaHES} \\
    \toprule
        \multirow{2}{*}{Spotlight} & 0.54 & -0.79 & -0.79 & 0.54 & 0.54 & 0.97 & \textbf{0.99}\\
        & $\pm$ 0.000 & $\pm$ 0.107 & $\pm$ 0.107 & $\pm$ 0.000 & $\pm$ 0.000 & $\pm$ 0.016 & $\mathbf{\pm}$ \textbf{0.005}\\
    \midrule
        \multirow{2}{*}{Euclidean} & 0.54 & -0.57 & -0.62 & 0.54 & 0.21 & 0.97 & \textbf{0.99}\\
        & $\pm$ 0.000 & $\pm$ 0.492 & $\pm$ 0.423 & $\pm$ 0.000 & $\pm$ 0.258 & $\pm$ 0.029 & $\mathbf{\pm}$ \textbf{0.011}\\
    \bottomrule
    \end{tabular}
\end{table}

\section{Details of Protein Sequence Design Experiments}
\label{sec:protein}
\subsection{Oracle Goodness of Fit}
We assess the Oracle model's performance with increasing training data to gauge how well our semi-synthetic setting approximates real data. The Oracle is a Bayesian linear regression model that takes protein embeddings from various large language models, ranging from Llama-2 7B (a general-purpose model) to ESM-2 3B (a protein-specific model). Our results show that the Gemma 7B embeddings yield the best regression performance, leading us to use Gemma 7B in subsequent experiments.
\begin{figure}[!htb]
    \centering
    \includegraphics[width=0.8\textwidth]{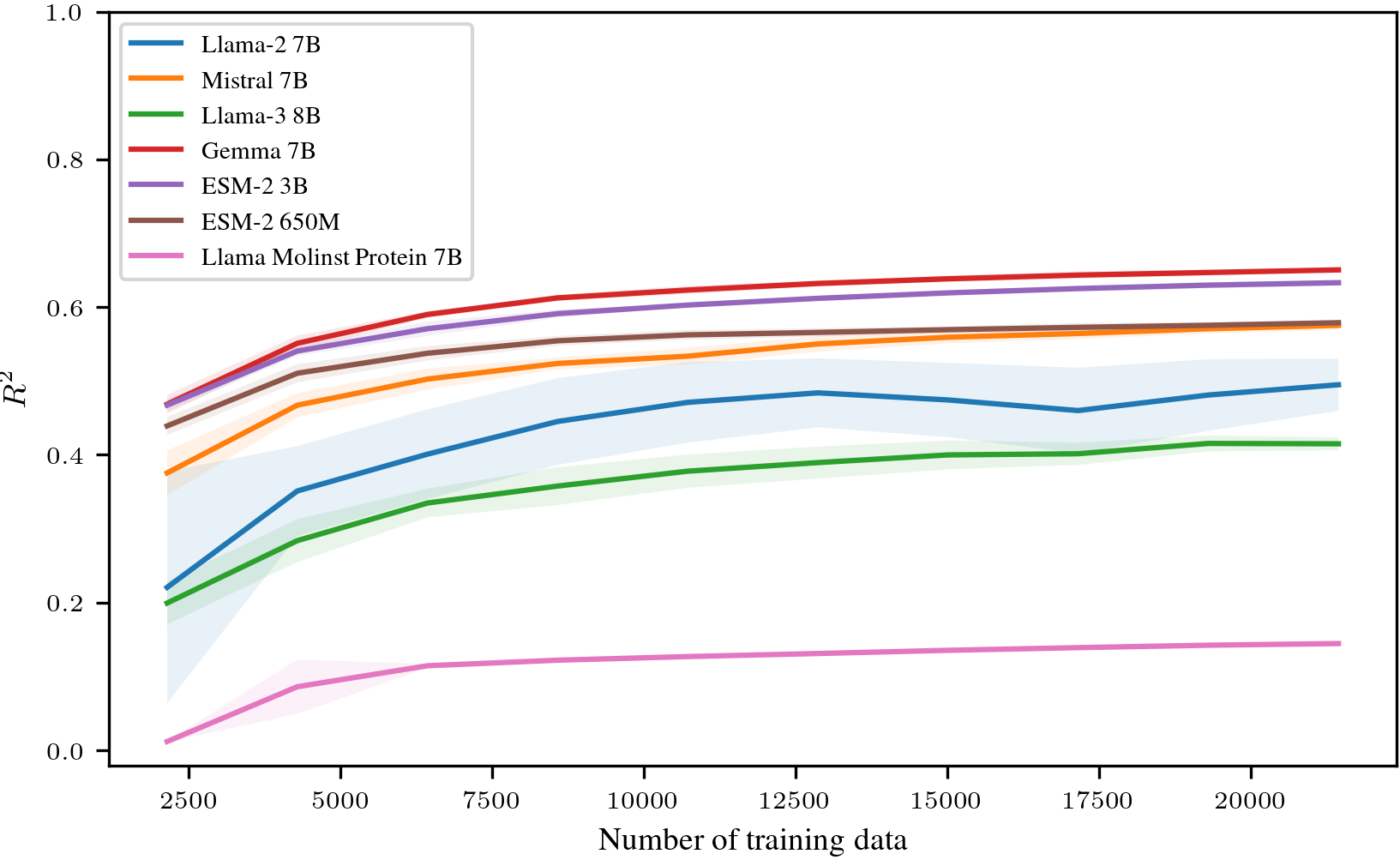}
    \caption{Coefficient of determination on the test set as a function of the number of training data. The \(R^2\) metric is used to evaluate the performance of embedding protein sequences.}
    \label{fig:goodness_of_fit}
\end{figure}

\subsection{Modeling Protein Sequence Design with Natural Language}
We employ the instruction-finetuned Llama-3.2 3B model as our variational network. The model is available online at \url{https://huggingface.co/meta-llama/Llama-3.2-3B-Instruct}. We frame the protein design process as a dialogue to leverage the model's conversational capabilities. Specifically, we prompt the model to generate the next protein sequences based on previously observed protein and their fluorescent level. The prompts we used are outlined below.

System prompt:
\begin{lstlisting}
    You are a helpful assistant who works in a protein engineering lab. We are trying to edit a given protein by a sequence of 1-step protein editing, known as mutation. You need to use your knowledge to help me propose suitable protein editing. Going from an initial protein to an optimal one can take many steps.
\end{lstlisting}

First prompt:
\begin{lstlisting}
    Edit 1 amino acid in the below protein sequence to create a new protein with higher fluorescence. The amino acid must be in set {D, E}. Protein sequence: {starting_protein}
\end{lstlisting}

Feedback prompt:
\begin{lstlisting}
    Fluorescence level of the above protein: {fluorescence_level} Based on the above protein sequence and its fluorescence value, edit 1 amino acid to achieve higher fluorescence. You must only return the modified protein sequence and nothing else. Modified protein sequence: 
\end{lstlisting}

\subsection{Supervised fine-tuning Process}
Before starting the BO process, we conduct supervised fine-tuning (SFT) on the variational network to adapt it to the protein design task. We generate a dataset for SFT training consisting of 100 dialogues, each containing $L$ rounds corresponding to the number of lookahead steps. The proteins in each dialogue are created by either randomly mutating or retaining the previous protein. The fine-tuning hyperparameters are provided below.

\begin{table}[htb!]
    \centering
    \caption{SFT Hyperparameters}
        \begin{tabular}{ll}
        \toprule
        \textbf{Hyperparameter}              & \textbf{Value}       \\ 
        \midrule
        Learning rate                   & $10^{-4}$            \\ 
        Epochs                          & $3.3$                \\ 
        Batch size                      & $4$                  \\ 
        Learning rate warmup ratio      & $0.1$                \\ 
        Learning rate schedule          & Cosine               \\ 
        LoRA $\alpha$                   & $32$                 \\ 
        LoRA $r$                        & $16$                 \\ 
        LoRA dropout                    & $0.1$                \\ 
        LoRA target modules             & q\_proj, v\_proj     \\ 
        \bottomrule
        \end{tabular}
    \label{tab:parameters}
\end{table}

\subsection{Nonmyopic Bayesian Optimization as Multi-turn Proximal Policy Optimization}
\begin{figure}
    \centering
    \includegraphics[width=\linewidth]{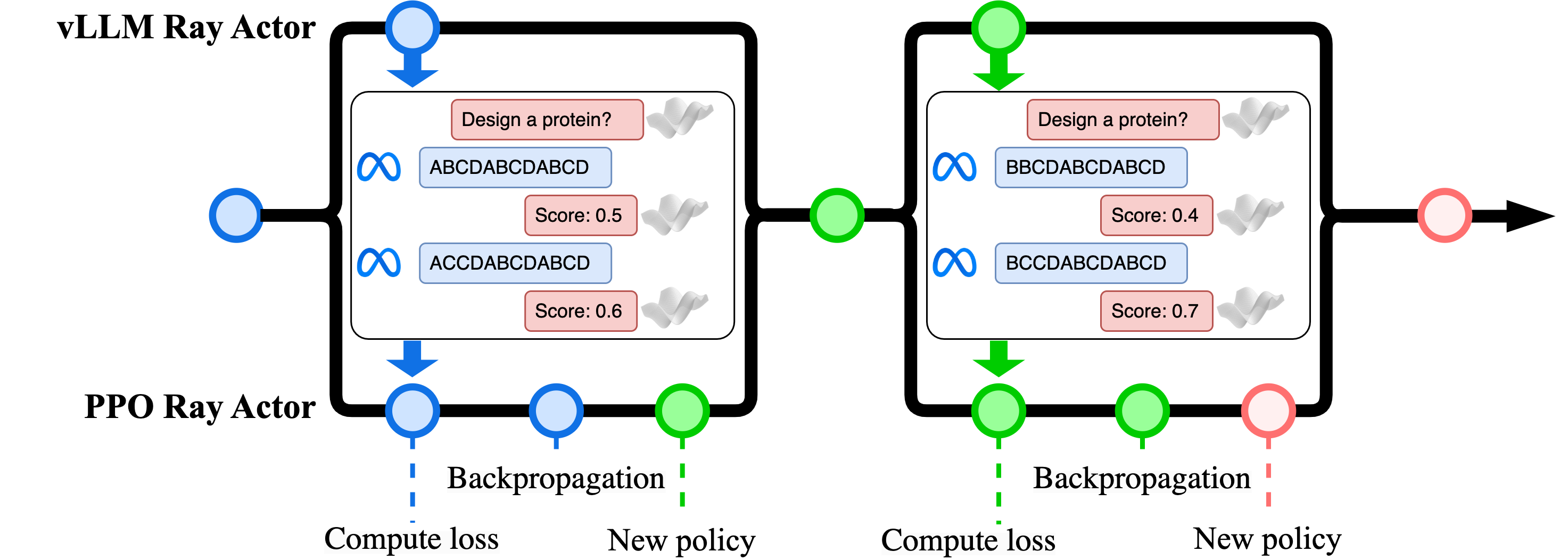}
    \caption{The designed PPO pipeline with vLLM using Ray involves the vLLM actor generating multi-turn protein refinements and calling a surrogate model to compute reward scores. These scores, along with the generated data, are passed to the PPO actor to compute loss, backpropagate, and update the LLM weights. The updated weights from the PPO actor are then synced to the vLLM actor for the next generation.}
    \label{fig:vllm_ray}
\end{figure}

Proximal Policy Optimization (PPO)~\cite{schulman2017proximal} is typically used to fine-tune language models for single-turn conversations, where the model responds once to a prompt without considering future turns. However, our approach requires the model to think ahead and generate multiple future queries (in this case, protein sequences) over several turns. To address this, we modify existing PPO frameworks to handle multiturn conversations, allowing the model to generate and optimize future sequences during training. We also use vLLM~\cite{kwon2023efficient}, a system designed to improve the speed and efficiency of inference (i.e., generating outputs from the model). However, vLLM is an inference engine and cannot be used directly for training~\cite{kwon2023efficient}. To overcome this, after each step of updating the model during training (called a gradient step), we transfer the updated model's weights (parameters) to the vLLM system. This allows us to use vLLM for faster generation of outputs, leading to more efficient training. Additionally, running vLLM with PPO simultaneously on multiple GPUs is challenging for large language models due to differences in how vLLM and PPO handle GPU VRAM. To address this, we employ Ray~\cite{Moritz2018Ray} to isolate the GPU environments for vLLM and PPO, creating a pipeline that enables efficient weight synchronization between them. Figure~\ref{fig:vllm_ray} illustrates the process of weight syncing between PPO and vLLM.

In the PPO training process, we calculate a final reward for each dialogue using a function \(\ell\). This function varies depending on the acquisition method being used (e.g., expected improvement or simple regret). Once the reward is computed, it is adjusted, or ``discounted,'' for each individual turn in the dialogue. This means that actions taken earlier in the conversation get less reward compared to later actions. We then use this discounted reward as feedback to update the model during PPO training. By doing this, we extend the single-turn PPO framework, which normally handles one response at a time, to work for our multiturn conversation data. The hyperparameters used for fine-tuning PPO are provided below.

\begin{table}[hbt!]
    \centering
    \caption{PPO and Rollout Hyperparameters}
    \begin{tabular}{ll}
        \toprule
        \textbf{Hyperparameter}              & \textbf{Value}       \\ 
        \midrule
        Learning rate                   & $10^{-4}$            \\ 
        Epochs                          & $64$                 \\ 
        Batch size                      & $1$                  \\ 
        Learning rate warmup ratio      & $0.1$                \\ 
        Learning rate schedule          & Cosine               \\ 
        LoRA $\alpha$                   & $256$                \\ 
        LoRA $r$                        & $128$                \\ 
        LoRA dropout                    & $0.1$                \\ 
        LoRA target modules             & q\_proj, v\_proj     \\ 
        Maximal rollout retry           & $32$                 \\ 
        Discount reward factor          & $0.95$               \\ 
        \bottomrule
    \end{tabular}
    \label{tab:new_parameters}
\end{table}

\section{Ablation Study on Protein Design Experiments}
\label{sec:protein_exp_ablation}

We conduct an ablation study using two different starting proteins and two distinct synthetic functions $g(x)$ to construct diverse protein spaces, enabling a systematic analysis of their characteristics. The two synthetic functions are defined as follows:  

\begin{equation}
    \begin{aligned}
        g_1(x) &= -0.005(d - 0.5)(d - 5)(d - 8)(d - 13.4),\\
        g_2(x) &= -e^{-0.7 \cdot \sqrt{0.5 \cdot d^2}} - e^{0.5 \cdot \cos(0.4 \pi d)} + e + 0.3.
    \end{aligned}
\end{equation}

\begin{figure}[!htb]
     \centering
     \hfill
     \begin{subfigure}[b]{0.45\textwidth}
         \centering
         \includegraphics[width=\textwidth]{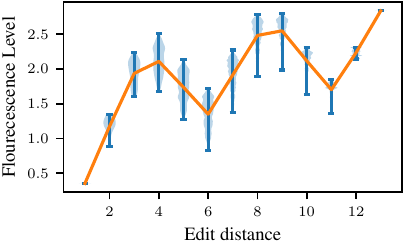}
         \caption{SP\#1, $g_2$}
         \label{fig:sp1_g2}
     \end{subfigure}
     \hfill
     \begin{subfigure}[b]{0.45\textwidth}
         \centering
         \includegraphics[width=\textwidth]{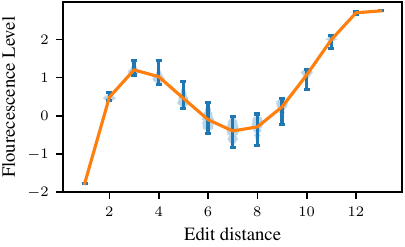}
         \caption{SP\#2, $g_1$}
         \label{fig:sp2_g1}
     \end{subfigure}
     \caption{Ablation of protein spaces with varying starting proteins and synthetic functions}
    \label{fig:ablation_protein_space}
\end{figure}

Here, $g_1(x)$ represents a polynomial function that introduces one local maxima across the input space, while $g_2(x)$ is a more complex one with two local maxima. Visualizations of the resulting protein spaces, derived from the combination of starting proteins and synthetic functions, are shown in Figure~\ref{fig:ablation_protein_space}.

\begin{table}[!htb]
    \caption{Protein Space Constraints}
    \label{tab:protein_space_constraints}
    \centering
    \begin{tabular}{m{0.3cm} m{8cm}  m{3cm}  m{1cm}}
    \toprule
        \textbf{No.} & \textbf{Starting protein} & \textbf{Allowed positions} & \textbf{Allowed AAs} \\
    \midrule
\#1& \texttt{\seqsplit{SKGEELFTGVVPILVELGGDVNGHKFSVSGEGEGDATYGKLTLKFICTTGKLPVPWPTLVTTLSYGVQCFSRFPDHMKQHDFFKSAMPEGYVQERTIFSKDDGNYKTRAEVKFEGDELVNRIELKGIDFKEEENILGHKLEENYNSHNVYIMADDQKNGIKVNFKIRHNIEDDSVQLADHYQQNTPIGDEPVLLPDDHYLSTQSALSKDDNEDRDEMVLLEFVTAAGITHGMDELYK}}
& 116, 131, 132, 141, 154, 171, 172, 189, 196, 209, 212, 215
& \texttt{E, D} \\
\midrule
\#2 & \texttt{\seqsplit{SKPEELFTPVVGILVELDPDVNGHKFSVSGEGEPDATYGKLTLKFICTTGKLGVGWGTLVTTLSYGVQCFSRYPDHMKQHDFFKSAMPEGYVQERTIFFKDDGNYKTRAEVKFEPDTLVNRIELKGIVFKEDGNTLGHKLEYNYNSHNVYIMADEQKNGIKVNFKIRHNIEDGSVQLADHYQQNTPIPDGPVLLPDNHYLSTQSALSKDPNEKRDHMVLLEFVTAAGITHGMDELYK}}
&  2, 8, 11, 18, 33, 52, 54, 56, 114, 158, 187, 190
& \texttt{G, P} \\
    \bottomrule
    \end{tabular}
\end{table}

\begin{figure}[hbpt!]
    \centering
    \includegraphics[width=0.7\linewidth]{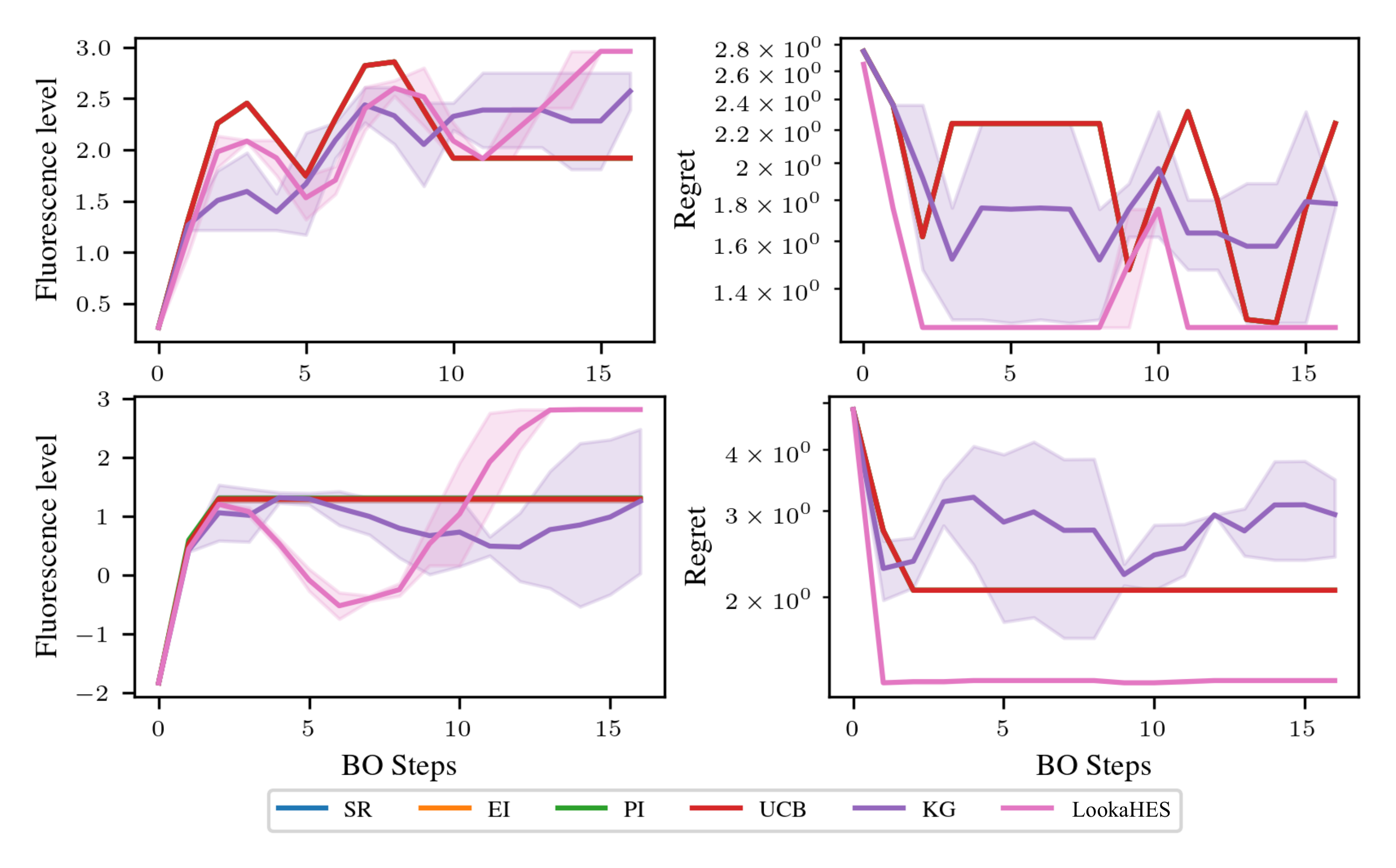}
    \caption{Fluorescence levels (left) and regret (right) observed over Bayesian Optimization (BO) steps. The top row shows results for experiments starting with protein \#1 and function $g_2$, while the bottom row corresponds to experiments starting with protein \#2 and function $g_1$.}
    \label{fig:m1f2_m2f1_yA}
\end{figure}

We present the results of additional experiments on protein design with the same starting protein with $g_2$ (Figure~\ref{fig:m1f2_m2f1_yA} top), and with a different starting protein with $g_1$ (Figure~\ref{fig:m1f2_m2f1_yA} bottom). These figures demonstrate that our proposed nonmyopic method outperforms other myopic baselines in various settings regardless of different starting proteins or synthetic value functions.

We visualize the designed proteins in the experiments of starting protein \#1 and $g_1$. We use ESMFold~\cite{lin2022language} to fold the designed proteins and PyMol~\cite{PyMOL} to visualize them. The visualizations are presented in Table~\ref{tab:designed_prots}.

\begin{table}[!hbt]
    \caption{Visualization of Designed Proteins. Edited Amino Acids Are Highlighted in {\color{red}Red}.}
    \label{tab:designed_prots}
    \centering
    \begin{tabular}{m{9cm} c}
        \toprule
        \textbf{Sequence} & \textbf{3D Structure} \\
        \midrule
        \textbf{Starting protein:} \texttt{\seqsplit{SKGEELFTGVVPILVELGGDVNGHKFSVSGEGEGDATYGKLTLKFICTTGKLPVPWPTLVTTLSYGVQCFSRFPDHMKQHDFFKSAMPEGYVQERTIFSKDDGNYKTRAEVKFEGDELVNRIELKGIDFKEEENILGHKLEENYNSHNVYIMADDQKNGIKVNFKIRHNIEDDSVQLADHYQQNTPIGDEPVLLPDDHYLSTQSALSKDDNEDRDEMVLLEFVTAAGITHGMDELYK}} & 
        \begin{minipage}{0.15\textwidth}
            \includegraphics[width=\linewidth]{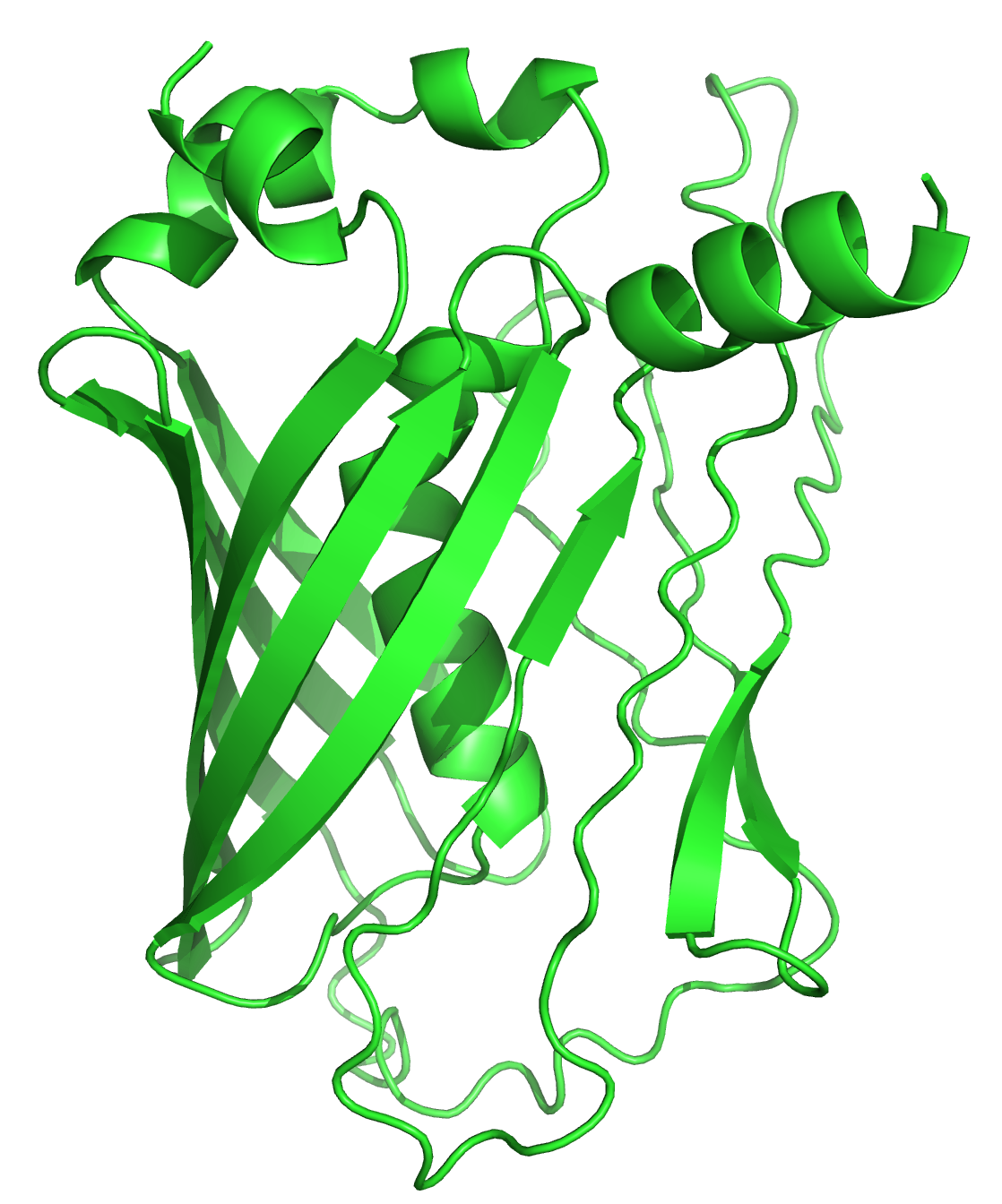}
        \end{minipage}\\
        
        \textbf{LookaHES - Optimal:} \texttt{\seqsplit{SKGEELFTGVVPILVELGGDVNGHKFSVSGEGEGDATYGKLTLKFICTTGKLPVPWPTLVTTLSYGVQCFSRFPDHMKQHDFFKSAMPEGYVQERTIFSKDDGNYKTRAEVKFEGD{\color{red}D}{\color{black}}LVNRIELKGIDFKE{\color{red}DD}{\color{black}}NILGHKLE{\color{red}D}{\color{black}}NYNSHNVYIMAD{\color{red}E}{\color{black}}QKNGIKVNFKIRHNIE{\color{red}EE}{\color{black}}SVQLADHYQQNTPIGD{\color{red}D}{\color{black}}PVLLPD{\color{red}E}{\color{black}}HYLSTQSALSKD{\color{red}E}{\color{black}}NE{\color{red}E}{\color{black}}RD{\color{red}D}{\color{black}}MVLLEFVTAAGITHGMDELYK}} & 
        \begin{minipage}{0.15\textwidth}
            \includegraphics[width=\linewidth]{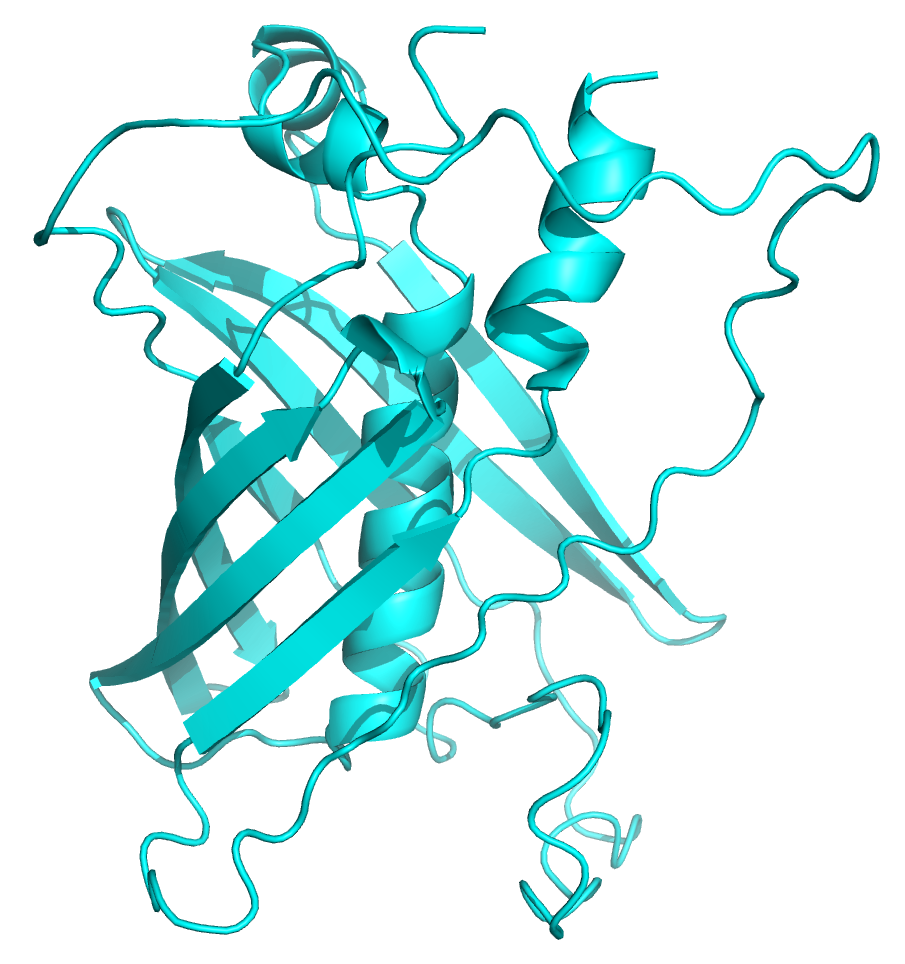}
        \end{minipage}\\
        
        \textbf{SR:} \texttt{\seqsplit{SKGEELFTGVVPILVELGGDVNGHKFSVSGEGEGDATYGKLTLKFICTTGKLPVPWPTLVTTLSYGVQCFSRFPDHMKQHDFFKSAMPEGYVQERTIFSKDDGNYKTRAEVKFEGDELVNRIELKGIDFKEE{\color{red}D}{\color{black}}NILGHKLEENYNSHNVYIMADDQKNGIKVNFKIRHNIEDDSVQLADHYQQNTPIGD{\color{red}D}{\color{black}}PVLLPDDHYLSTQSALSKDDNEDRDEMVLLEFVTAAGITHGMDELYK}} & 
        \begin{minipage}{0.15\textwidth}
            \includegraphics[width=\linewidth]{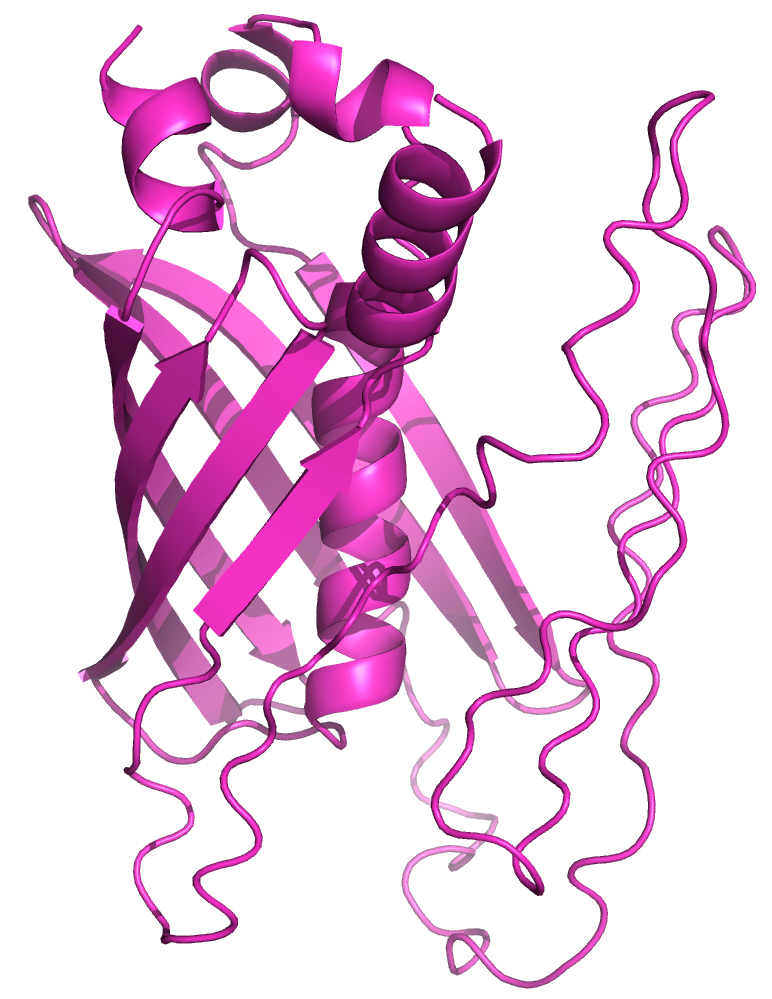}
        \end{minipage}\\
        
        \textbf{EI, PI, UCB:} \texttt{\seqsplit{SKGEELFTGVVPILVELGGDVNGHKFSVSGEGEGDATYGKLTLKFICTTGKLPVPWPTLVTTLSYGVQCFSRFPDHMKQHDFFKSAMPEGYVQERTIFSKDDGNYKTRAEVKFEGDELVNRIELKGIDFKE{\color{red}D}{\color{black}}ENILGHKLEENYNSHNVYIMADDQKNGIKVNFKIRHNIEDDSVQLADHYQQNTPIGDEPVLLPDDHYLSTQSALSKD{\color{red}E}{\color{black}}NEDRDEMVLLEFVTAAGITHGMDELYK}} & 
        \begin{minipage}{0.15\textwidth}
            \includegraphics[width=\linewidth]{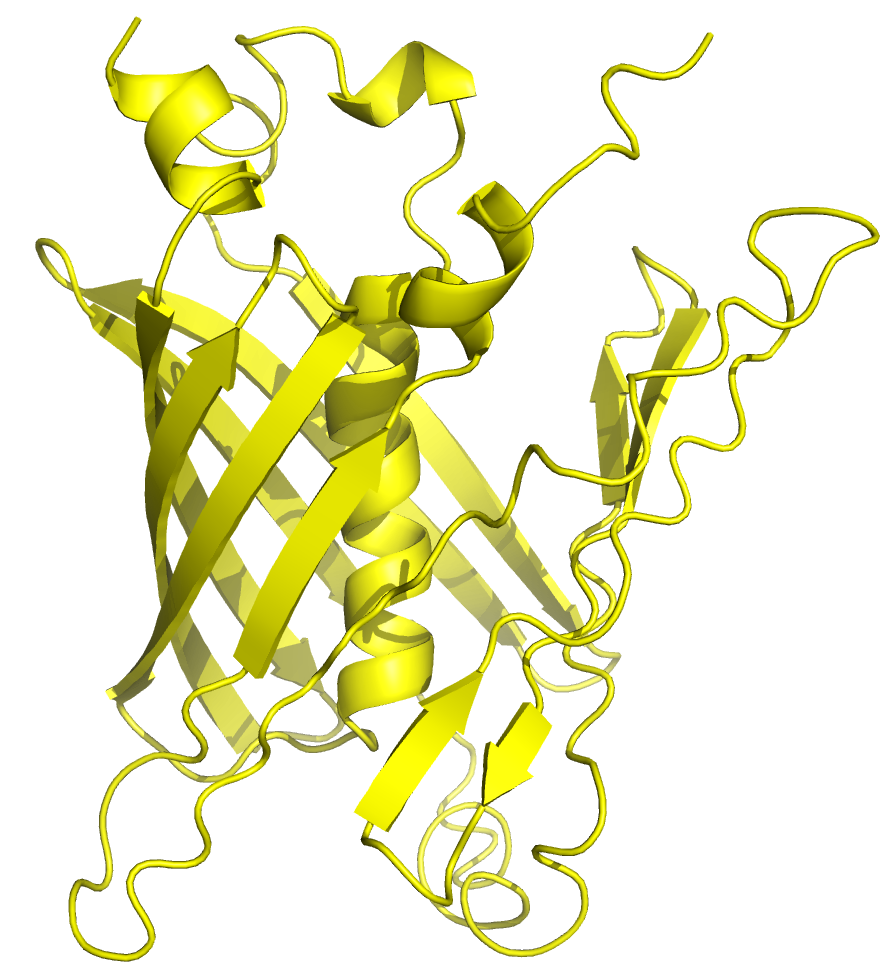}
        \end{minipage}\\
        
        \textbf{KG:} \texttt{\seqsplit{SKGEELFTGVVPILVELGGDVNGHKFSVSGEGEGDATYGKLTLKFICTTGKLPVPWPTLVTTLSYGVQCFSRFPDHMKQHDFFKSAMPEGYVQERTIFSKDDGNYKTRAEVKFEGDELVNRIELKGIDFKEEENILGHKLEENYNSHNVYIMADDQKNGIKVNFKIRHNIEDDSVQLADHYQQNTPIGDEPVLLPDDHYLSTQSALSKDDNE{\color{red}E}{\color{black}}RDEMVLLEFVTAAGITHGMDELYK}} & 
        \begin{minipage}{0.15\textwidth}
            \includegraphics[width=\linewidth]{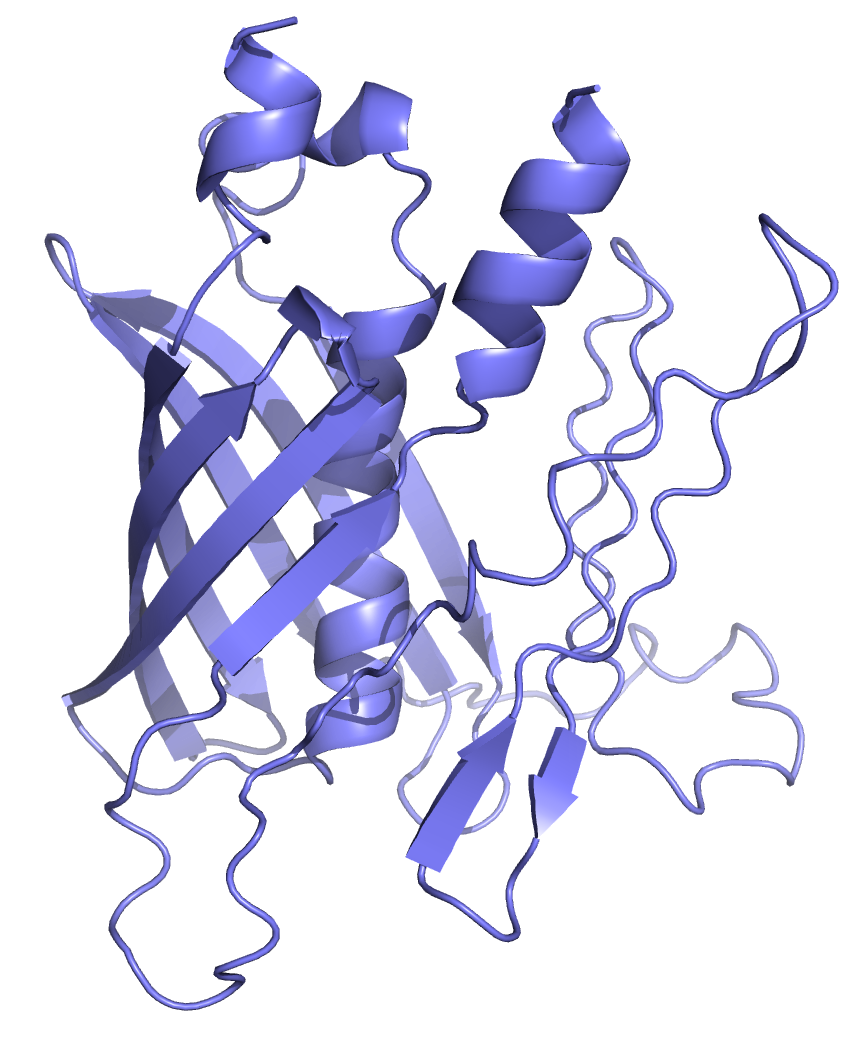}
        \end{minipage}\\
        \midrule
    \end{tabular}
\end{table}

%% file: sections/3.related_work.tex
\section{Related Works}
\label{further_related_work}
\paragraph{Nonmyopic Bayesian Optimization in the Dynamic Cost Setting}
Nonmyopic BO has been extensively explored in prior works~\cite{2010_Osborne, 2016_Gonzalez, 2019_Wu, 2020_Jiang, 2020_Lee, 2021_Lee, budgetedBO, folch2022snake, 2023_Belakaria, oneshottree}. These studies focus on converting a nested, multi-step planning problem into a single, high-dimensional optimization problem that can be solved efficiently with quasi-Monte Carlo sampling and gradient-based optimization. The advantage of having a lookahead mechanism is enlarging the receptive field—the area the decision maker can see prior to making a decision (Figure~\ref{fig:main_fig}). This approach has gained traction in cost-aware and budget-constrained BO~\cite{budgetedBO, 2021_Lee}, where nonmyopic planning is crucial. However, a notable challenge with this methodology is its limited scalability when extending the lookahead horizon, primarily due to the exponential increase in the number of decision variables. Our approach introduces a novel combination of Thompson sampling~\cite{Thompson1933ONTL} with the extensive generalization capabilities of a variational network, significantly enhancing the computational efficiency of nonmyopic BO. Utilizing neural networks for variational inference and optimization is not a new concept~\cite{kingma2022autoencodingvariationalbayes, amos2023tutorial}. For instance, Deep Adaptive Design~\cite{Foster2021DeepAD, ivanova2021implicit} is a parallel line of research from the Bayesian optimal experimental design literature, which has a related but distinct objective to reduce the uncertainty of model parameters as opposed to the global optimization objective in BO. The authors concentrate on reducing the computational demands during deployment and determining the most informative experimental designs by upfront offline optimization of a neural network to amortize the design cost. Our approach diverges by aligning more closely with the principles of online Bayesian optimization. Here, the primary objective extends beyond mere information acquisition to encompass the pursuit of global optimization. Our approach, which employs adaptive decision-making through online policy optimization, could be more robust than offline methods, particularly when the approximation of the reward function changes significantly between queries. This robustness arises because the online policy is updated at each BO step while the offline methods rely on transferring knowledge from learned offline data~\cite{pmlr-v235-nguyen-tang24a}. The distinctive feature of our study is implementing a pathwise, Thompson sampling-based nonmyopic acquisition function, which significantly reduces the computational cost of the iterative posterior sampling approach in~\cite{oneshottree}. Additionally, we present detailed comparisons of related works on different cost structures in Appendix~\ref{sec:cost_taxonomy}.

\paragraph{Variational Policy Optimization in Complex Action Spaces} In many real-world applications, decision-makers must take actions that are complex and subject to semantic constraints. Semantic constraints refer to rules or relationships that restrict the set of valid actions based on their meanings, dependencies, or contextual appropriateness. For example, in biological sequence design, semantic constraints may ensure that the mutated sequence is valid and does not unfold the protein (i.e., protein denaturalization). Recent RL research has addressed environments with such actions, which are challenging due to two main reasons: (i) the large number of potential actions~\cite{hubert2021learningplanningcomplexaction,zhang2024controllinglargelanguagemodelbased}, and (ii) the complex semantics~\cite{Thomas2023Grounding} underlying each action, making them difficult to capture. Recent studies have shown that modern LLMs can effectively model semantic actions and be fine-tuned with feedback from the environment~\cite{zhu2024languagemodelsinferaction, zhang2024controllinglargelanguagemodelbased,zhuang2024toolchain,hazra2024saycanpay}. Several papers demonstrate that using LLMs as policy models in reinforcement learning leads to better outcomes~\cite{palo2023towards, zhuang2024toolchain, hazra2024saycanpay}. In the field of NLP, a chatbot such as ChatGPT can be viewed as a decision-making process where the underlying LLM must understand user questions or requests to provide appropriate responses. These actions are complex and semantically rich, as even a single word can alter the meaning of a sentence. Consequently, RL methods like proximal policy optimization~\cite{schulman2017proximal} have been applied to refine the abilities of language models. 

\paragraph{Multi-turn Training Framework for LLMs} Multi-turn conversations have been shown to be more effective for managing entire dialogues~\cite{zhou2024archertraininglanguagemodel}. This approach can be viewed as a nonmyopic RL method that trains LLMs to achieve better conversational outcomes. Unfortunately, current RL training frameworks for LLMs, such as TRL~\cite{trl}, OpenRLHF~\cite{hu2024openrlhf}, LlamaFactory~\cite{zheng2024llamafactory}, and Nemo~\cite{HarperNemo}, primarily focus on single-turn conversations. As a result, they are not suited for multi-turn conversation training. When using these frameworks, multi-turn conversations must be divided into individual single turns, which limits the LLM's ability to manage the overall outcome of a conversation effectively.

%% file: sections/9.impact.tex
\section{Impact Statement}
This work presents a novel method for nonmyopic BO in dynamic cost settings, utilizing neural network policies to address long-horizon sequential decision-making. The proposed algorithm is computationally efficient and improves decision-making under uncertainty, with applications in complex scenarios like protein sequence design. Ethical implications include applications in healthcare (e.g., personalized treatment plans) and education (e.g., adaptive learning technologies), where effective and equitable resource allocation is critical. Mitigating biases and ensuring transparency will be essential to prevent systemic inequalities. This research advances machine learning for complex decision-making while emphasizing fairness and ongoing evaluation to ensure ethical deployment.